\documentclass[final,5p,times,twocolumn,authoryear]{elsarticle}

\usepackage{amssymb}
\usepackage{amsmath}

\usepackage{url}
\usepackage{adjustbox}
\usepackage{setspace}
\usepackage{array}
\usepackage{booktabs}
\usepackage{subcaption} 
\usepackage{calc}
\usepackage{graphicx}
\usepackage{multirow}
\usepackage{tabularx}
\usepackage{hyperref}

\journal{Medical Image Analysis}

\begin{document}

\begin{frontmatter}



\title{Latent Diffusion Autoencoders: Toward Efficient and Meaningful Unsupervised Representation Learning in Medical Imaging – A Case Study on Alzheimer’s Disease} 


\author[1,2]{Gabriele Lozupone\corref{main}\corref{cor1}} 
\ead{gabriele.lozupone@unicas.it}
\cortext[main]{Main contributor.}
\cortext[cor1]{Corresponding author:}
\author[1]{Alessandro Bria} 
\author[1]{Francesco Fontanella} 
\author[3]{Frederick J.A. Meijer} 
\author[1]{Claudio De Stefano} 
\author[2]{and Henkjan Huisman} 
\author[]{for the Alzheimer's Disease Neuroimaging Initiative\corref{fn1}} 

\affiliation[1]{organization={Department of Electrical and Information Engineering (DIEI), University of Cassino and Southern Lazio}, 
            addressline={Via G. Di Biasio 43}, 
            city={Cassino},
            postcode={03043}, 
            state={FR},
            country={Italy}}

\affiliation[2]{organization={Diagnostic Image Analysis Group, Radboud University Medical Center},
addressline={Geert Grooteplein 10},
city={Nijmegen}, 
postcode={6500HB},
country={Netherlands}
}

\affiliation[3]{organization={Department of Medical Imaging, Radboud University Medical Center},
addressline={Geert Grooteplein 10},
city={Nijmegen}, 
postcode={6500HB},
country={Netherlands}
}

\fntext[fn1]{*Data used in preparation of this article were obtained from the Alzheimer's Disease Neuroimaging Initiative (ADNI) database (adni.loni.usc.edu). As such, the investigators within the ADNI contributed to the design and implementation of ADNI and/or provided data but did not participate in the analysis or writing of this report. A complete listing of ADNI investigators can be found at: \url{http://adni.loni.usc.edu/wp-content/uploads/how_to_apply/ADNI_Acknowledgement_List.pdf}.}

\begin{abstract}

This study presents Latent Diffusion Autoencoder (LDAE), a novel encoder-decoder diffusion-based framework for efficient and meaningful unsupervised learning in medical imaging, focusing on Alzheimer’s disease (AD) using brain MR from the ADNI database as a case study. Unlike conventional diffusion autoencoders operating in image space, LDAE applies the diffusion process in a compressed latent representation, improving computational efficiency and making 3D medical imaging representation learning tractable. To validate the proposed approach, we explore two key hypotheses: (i) LDAE effectively captures meaningful semantic representations on 3D brain MR associated with AD and ageing, and (ii) LDAE achieves high-quality image generation and reconstruction while being computationally efficient. Experimental results support both hypotheses: (i) linear-probe evaluations demonstrate promising diagnostic performance for AD (ROC-AUC: 90\%, ACC: 84\%) and age prediction (MAE: 4.1 years, RMSE: 5.2 years); (ii) the learned semantic representations enable attribute manipulation, yielding anatomically plausible modifications; (iii) semantic interpolation experiments show strong reconstruction of missing scans, with SSIM of 0.969 (MSE: 0.0019) for a 6-month gap. Even for longer gaps (24 months), the model maintains robust performance (SSIM $>$ 0.93, MSE $<$ 0.004), indicating an ability to capture temporal progression trends; (iv) compared to conventional diffusion autoencoders, LDAE significantly increases inference throughput (20x faster) while also enhancing reconstruction quality. These findings position LDAE as a promising framework for scalable medical imaging applications, with the potential to serve as a foundation model for medical image analysis. Code available at \url{https://github.com/GabrieleLozupone/LDAE}

\end{abstract}



\begin{keyword}
Alzheimer's disease
\sep Diffusion Models 
\sep Foundation Models
\sep Representation Learning



\end{keyword}

\end{frontmatter}

\section{Introduction}
\label{sec:intro}
Recently, Diffusion Probabilistic Models (DPMs) have shown remarkable performance in image synthesis and dataset distribution modelling \citep{ddpm_dhariwal2021diffusion, diff_nichol2021glide}. The DPMs' stable training process allowed to achieve state-of-the-art sample quality surpassing deep generative models like generative adversarial networks (GANs) \citep{gan_goodfellow2014generative} and variational autoencoders (VAEs) \citep{vae_kingma2013auto, vae_rezende2014stochastic}. Most of the existing literature explores DPMs synthesis and editing capability, with limited focus on their representational capability \citep{diff_rl_xiang2023denoising, diff_rl_abstreiter2021diffusion, soda_hudson2024soda}. One of the first works to propose diffusion-based approaches for representation learning was \cite{dae_preechakul2022diffusion}. The authors introduced Diffusion Autoencoders (DAEs) to produce a meaningful and decodable representation by means of an encoder to capture high-level semantics and a diffusion-based decoder for the low-level stochastic variations. For the first time, a Diffusion-based model surpassed GANs in feature disentanglement and enabled image attribute manipulation, preserving image quality and training stability. Traditional DPMs can act as encoder-decoder by converting an input image \(x_0\) into a spatial latent variable \(x_T\) by running the diffusion process backwards. The latent representation, however, lacks semantics and properties such as disentanglement, compactness, and the ability to perform semantic interpolation. In contrast, DAEs ensure that representations are both compact and disentangled, while also facilitating meaningful linear interpolation in the latent space. This makes the DAE approach a strong candidate for structured semantic learning. Subsequently, some studies investigated different representation learning strategies based on encoder-decoder architectures like \cite{pdae_zhang2022unsupervised} that explored DAEs representation learning from pretrained DPMs (PDAE) and the more recent SODA architecture \citep{soda_hudson2024soda} that introduces layer modulation to improve semantic attribute disentanglement further in the semantic space. Representation learning with diffusion models has received limited attention overall and even less in medical imaging. In 3D MRI brain domain diffusion-based approaches are mainly proposed for unconditional and conditional generation \citep{brain_diff_peng2023generating, brain_diff_pinaya2022brain} and for disease progression prediction using pre-computed brain anatomical features and MRI sequences \citep{brain_diff_longitudinal_yoon2023sadm, brain_diff_longitudinal_puglisi2024enhancing}. Therefore, unsupervised representation learning using diffusion models to learn a general semantic representation that captures the complex 3D brain anatomical structure remain an unexplored direction. 

\paragraph{Contributions}This work introduces Latent Diffusion Autoencoders (LDAE) as an efficient  framework for unsupervised and meaningful representation learning. The proposed approach builds upon principles established in Diffusion Autoencoders (DAE) \citep{dae_preechakul2022diffusion}, Pretrained Diffusion Autoencoders (PDAE) \citep{pdae_zhang2022unsupervised}, and Latent Diffusion Models (LDMs) \citep{sd_rombach2022high}. Unlike conventional diffusion-based autoencoders that operate in the original image space, LDAE applies the diffusion process in a compressed latent space. This formulation  enhances computational efficiency and scalability, making diffusion-based representation learning tractable for 3D medical imaging.

\paragraph{Method}Our approach consists of three key stages: (i) a perceptual autoencoder (AE) that compresses high-dimensional MRI scans into a lower-dimensional latent space; (ii) pretraining a diffusion model on the compressed latent representations; and (iii) LDAE unsupervised representation learning with an encoder-decoder to fill the posterior mean gap, following the strategy introduced in PDAE \citep{pdae_zhang2022unsupervised}. To the best of our knowledge, this is the first latent diffusion autoencoder framework, demonstrating that meaningful semantic representations can be learned even when the diffusion process is performed in a compressed space.

\paragraph{Experiments and Results} 
To validate the effectiveness of the proposed LDAE, we conduct several experiments using longitudinal 3D brain MRI data from the Alzheimer’s Disease Neuroimaging
Initiative (ADNI) database. Our evaluation aims to validate two key hypotheses: (i) LDAE learns semantically rich representations that capture clinically relevant attributes related to Alzheimer’s disease (AD) and ageing consenting unsupervised representation learning via latent-space DAE, and (ii) LDAE offers high-quality generation and reconstruction improving computational efficiency over conventional DAE. For hypothesis (ii), we compare LDAE to voxel-space DAEs and demonstrate a significant improvement in efficiency, achieving a 20$\times$ speedup in inference time while surpassing reconstruction quality (SSIM: 0.962, MSE: 0.001, LPIPS: 0.076). To assess hypothesis (i), we perform linear-probe evaluations on the learned semantic embeddings. LDAE achieves good performance in downstream tasks such as AD diagnosis (ROC-AUC: 89.48\%, Accuracy: 83.65\%) and age prediction (MAE: 4.16 years, RMSE: 5.23 years), indicating that the learned latent codes encode clinically meaningful information. We further validate semantic interpretability through latent attribute manipulation, which consent brain MR anatomical structures alteration related to disease and age progression. Finally, semantic and stochastic interpolation experiments show LDAE’s capacity to predict missing intermediate scans in longitudinal series, achieving robust performance even at longer temporal gaps, supporting its ability to capture the temporal trajectory of neurodegeneration (e.g., SSIM: 0.97 for 6-month intervals and SSIM $>$ 0.93 for 24-month gaps; SSIM is computed between the reconstructions done by the autoencoder to isolate the interpolation quality from the upper-bound limitation due to the compression model).


The remainder of the paper is organized as follows: Section  \ref{sec:background} introduces DPMs and background concepts utilized in the work, Section \ref{sec:methods} describes in detail the multi-stage approach investigated in this manuscript. Section \ref{sec:experiments} presents the experimental settings, and Section \ref{sec:results} the generation, manipulation, interpolation and downstream tasks results. Discussions of the findings and conclusions are provided in Section \ref{sec:conclusions}.

\begin{figure*}[!ht]
\centering
\includegraphics[width=\textwidth]{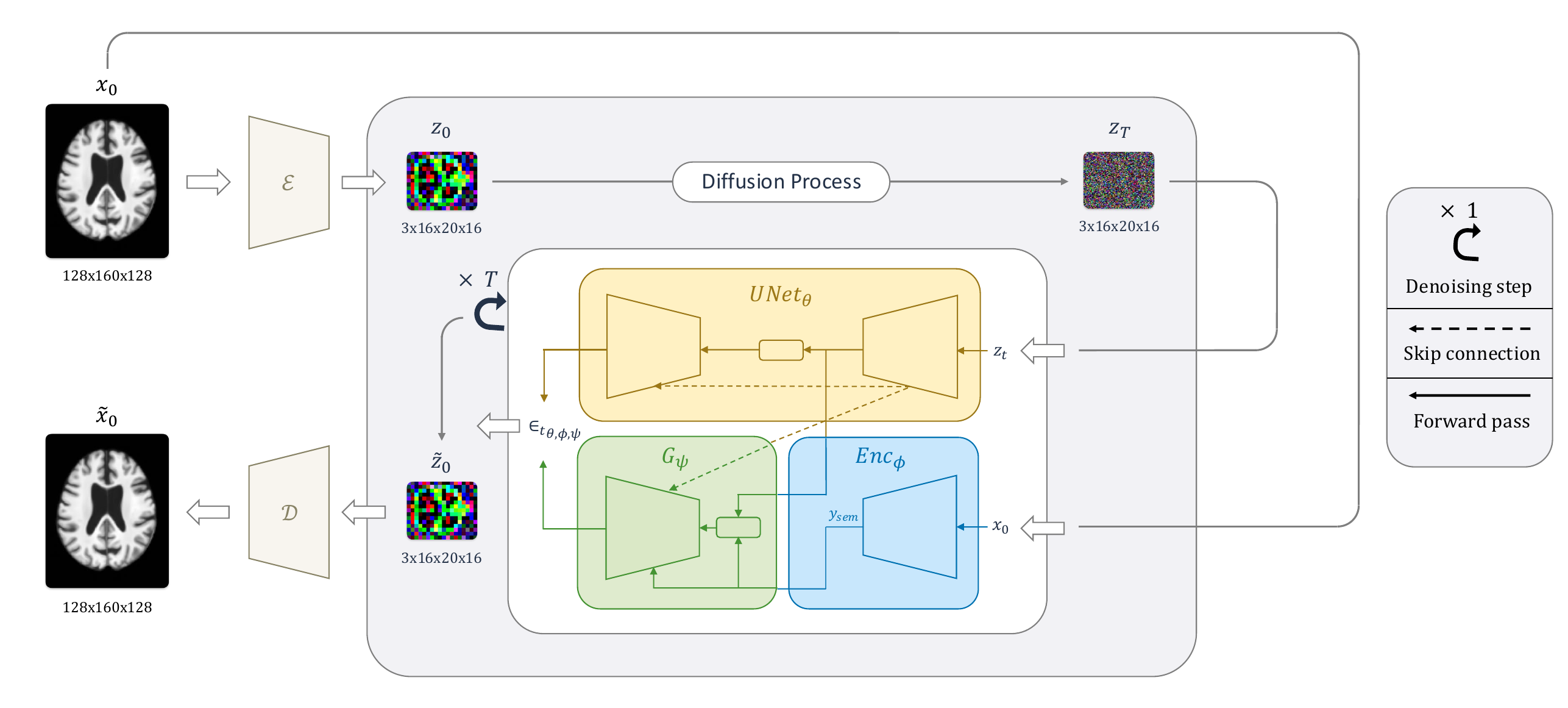}
\caption{Overview of the proposed 3D LDAE framework for unsupervised representation learning in brain medical imaging. The framework consists of three key components: (i) a \textbf{compression model} (\(\mathcal{E}\) and \(\mathcal{D}\)), which encodes high-dimensional MRI brain scans (\(\mathbf{x}_0\)) into a lower-dimensional latent representation (\(\mathbf{z}_0\)), facilitating efficient processing; (ii) a \textbf{Latent Diffusion Model (LDM)}, trained to learn the distribution of the compressed representations through a diffusion process, progressively transforming \(\mathbf{z}_0\) into \(\mathbf{z}_T\) and vice versa via a U-Net-based denoising network (\(\text{UNet}_\theta\)); and (iii) the \textbf{semantic encoder-decoder model} (\(G_\psi\) and \(Enc_\phi\)), which learns a meaningful latent representation (\(\mathbf{y}_{sem}\)) from the input scan and utilizes it to guide the reverse diffusion process via a gradient estimator (\(G_\psi\)). This approach enables structured semantic learning, facilitating interpretable image synthesis, counterfactual generation, and disease-specific attribute disentanglement.}
\label{fig:approach_overview}
\end{figure*}

\section{Background}
\label{sec:background} 
DPMs are generative models that model a target distribution by learning a denoising process at varying noise levels \citep{sohl2015deep}. This concept is inspired by nonequilibrium thermodynamics, in which a physical system starts from a structured, low-entropy state that is gradually "diffused" or driven toward a more disordered, high-entropy equilibrium state over time. In principle, the system can be steered back toward a more ordered configuration, although this typically requires precise control and information about the underlying dynamics. In diffusion-based generative models, we begin with real data and then apply a stochastic “diffusion” of noise step-by-step. Each step slightly corrupts the data by adding Gaussian noise to arrive at a highly noisy, nearly featureless distribution that is mathematically close to a pure Gaussian distribution \(\mathcal{N}(\mathbf{0}, \mathbf{I})\).
\subsection{Denoising Diffusion Probabilistic Models}
Denoising Diffusion Probabilist Models (DDPMs) proposed in \cite{ho2020denoising} defined the diffusion process as a Markov chain that starts from the data distribution \(q(\mathbf{x}_0)\) and sequentially corrupts in \(T\) steps it to \(\mathcal{N}(\mathbf{0}, \mathbf{I})\) with Markov diffusion kernels \(q(\mathbf{x}_t | \mathbf{x}_{t-1})\). The kernels are defined by a fixed variance schedule \(\{ \beta_t \}_{t=1}^T\) where $\alpha_t = 1 - \beta_t$ and $\bar{\alpha}_t = \prod_{i=1}^{t} \alpha_i$. This formulation allows to directly sample \(\mathbf{x}_t\) from \(\mathbf{x}_0\) for arbitrary \(t\) with \(
q(\mathbf{x}_t | \mathbf{x}_0) = \mathcal{N}(\mathbf{x}_t ; \sqrt{\bar{\alpha}_t} x_0 , (1 - \bar{\alpha}_t) \mathbf{I})
\). The overall process can be expressed by:
\begin{equation}
\begin{aligned}
    q(\mathbf{x}_t | \mathbf{x}_{t-1}) &= \mathcal{N} \Big( \mathbf{x}_t ; \sqrt{1 - \beta_t} \mathbf{x}_{t-1}, \beta_t \mathbf{I} \Big), \\
    q(\mathbf{x}_{1:T} |\mathbf{x}_0) &= \prod_{t=1}^{T} q(\mathbf{x}_t | \mathbf{x}_{t-1}).
\end{aligned}
\label{eq:diffusion_process}
\end{equation} We are interested in learning the reverse process, i.e., the distribution \(p(\mathbf{x}_{t-1}|\mathbf{x}_t)\). As shown by \cite{sohl2015deep}, these probability functions are difficult to model unless the gap between \(t-1\) and \(t\) is infinitesimally small (\(t \to \infty\)). In practice, a sufficiently large \(T=1000\) is chosen and in such a case, a good approximation \(p_\theta(\mathbf{x}_{t-1}|\mathbf{x}_t)\) can be modeled as \(\mathcal{N}(\mu_\theta(\mathbf{x}_t, t), \sigma_t)\) in which parameters \(\theta\) can be learned with an UNet \citep{ho2020denoising}. The model is trained with the loss function: \(L_\epsilon=||\mathbf{\epsilon_\theta}(\mathbf{x}_t, t) - \epsilon||\), where \(\epsilon\) is the noise added to \(\mathbf{x}_0\) to obtain \(\mathbf{x}_t\). This is a simplified formulation of the variational lower bound on the marginal log-likelihood commonly used in DDPMs training \citep{dhariwal2021diffusion, nichol2021improved, dae_preechakul2022diffusion, song2020denoising}.

\subsection{Denoising Diffusion Implicit Models}
Denoising Diffusion Implicit Models (DDIMs) proposed in \citep{song2020denoising} introduce a non-Markovian forward process that, unlike standard DDPMs, offers increased flexibility, enabling faster inference with fewer steps. In particular, the latent variable $x_{t-1}$ can be derived from $x_t$ by leveraging $\epsilon_\theta$ from a pretrained DDPM as follows:
\begin{equation}
    \begin{aligned}
        \mathbf{x}_{t-1} = 
        &\sqrt{\bar{\alpha}_{t-1}} \left( \frac{\mathbf{x}_t - \sqrt{1 - \bar{\alpha}_t} \epsilon_{\theta_t}(\mathbf{x}_t, t)}
        {\sqrt{\bar{\alpha}_t}} \right) + \\
        &+ \left(\sqrt{1 - \bar{\alpha}_{t-1} - \sigma_t^2})\right) \epsilon_{\theta_t}(\mathbf{x}_t, t) + \sigma_t \epsilon_t.
    \end{aligned}
\label{eq:reverse_diffusion}
\end{equation}
where $\epsilon_t \sim \mathcal{N}(0, \mathbf{I})$, and $\sigma_t$ determines the degree of stochasticity in the forward process. By choosing $\sigma_t = 0$, the generative process will be fully deterministic, which is named implicit in DDIMs. The DDIM posterior distribution becomes:
\begin{equation}
    q(\mathbf{x}_{t-1} | \mathbf{x}_t, \mathbf{x}_0) = \mathcal{N} \left( 
    \sqrt{\bar{\alpha}_{t-1}} \mathbf{x}_0 + \sqrt{1 - \bar{\alpha}_{t-1}} \frac{\mathbf{x}_t - \sqrt{\bar{\alpha}_t} \mathbf{x}_0}{\sqrt{1 - \bar{\alpha}_t}}, 0
    \right).
\label{eq:posterior}
\end{equation}
while maintaining the DDPM marginal distribution: 
\begin{equation}
    q(\mathbf{x}_t | \mathbf{x}_0) = \mathcal{N}(\mathbf{x}_t ; \sqrt{\bar{\alpha}_t} \mathbf{x}_0 , (1 - \bar{\alpha}_t) \mathbf{I}).
\label{eq:marginal}
\end{equation}
Since $\sigma_t = 0$ implies a deterministic generative process, the DDIM can encode \(\mathbf{x}_0\) into a decodable noise map \(\mathbf{x}_t\). In \cite{dae_preechakul2022diffusion}, the authors show that this process yields an accurate reconstruction but \(\mathbf{x}_t\) lacks high-level semantics, not consenting a semantically-smooth interpolation between samples.

\subsection{Diffusion Autoencoders}
In the objective of obtaining a semantically meaningful latent code, the authors of DAE \citep{dae_preechakul2022diffusion} designed a conditional DDIM image decoder that approximate \(p(\mathbf{x}_{t-1}|\mathbf{x}_t, \mathbf{y}_{sem})\) in which \(\mathbf{y}_{sem}\) is a non-spatial vector of dimension \(d=512\) learned from a \textbf{semantic encoder} that maps \(\mathbf{x}_0\) into \(\mathbf{y}_{sem}=Enc_\phi(\mathbf{x}_0)\). The encoder and the DDIM decoder are trained in conjunction by optimizing: 
\begin{equation}
    L_{\epsilon} = \sum_{t=1}^{T} \mathbb{E}_{\mathbf{x}_0, \epsilon_t} \left[ \left\| \epsilon_{\theta} (\mathbf{x}_t, t, \mathbf{y}_{\text{sem}}) - \epsilon_t \right\|_2^2 \right].
\label{eq:diffusion_loss}
\end{equation}
with respect to \(\theta\) and \(\phi\) by conditioning an UNet with the \(Enc_\phi\) output using adaptive group-wise normalization (AdaGN) layers as proposed in \cite{ddpm_dhariwal2021diffusion}. By training the two models simultaneously, the encoder \(Enc_\phi\) is forced to learn as much information as possible to help the DDIM in the denoising process. The authors of PDAE \citep{pdae_zhang2022unsupervised} clarified this behaviour  by showing that there exists a gap between the posterior mean predicted by an unconditional DPMs \((\mu_\theta(\mathbf{x}_t,t))\) and the true one \((
\tilde{\mu}_t(\mathbf{x}_t, \mathbf{x}_0)
)\). The posterior mean gap is caused by an information loss that, in theory, can be recovered by conditioning on some \(\mathbf{y}\) that contain all information about \(\mathbf{x}_0\). By letting \(\mathbf{y}=\mathbf{y}_{sem}\) a learnable vector produced by an encoder, it will be forced to learn as much information as possible to fill the gap and consequently as much information as possible from \(\mathbf{x}_0\). Following these principles, it is possible to train a DAE from pretrained DPMs, achieving better training efficiency and stability \citep{pdae_zhang2022unsupervised}.

\subsection{Classifier-Guided Sampling Method}

The classifier-guided method allows to condition the generation of a DDPM towards some information, e.g. classes or prompts \citep{sohl2015deep, song2020score}. It consists of training a classifier $p_{\psi}(\mathbf{y}|\mathbf{x}_t)$ on noisy data. The gradient $\nabla_{\mathbf{x}_t} \log p_{\psi}(\mathbf{y}|\mathbf{x}_t)$ can then be leveraged to guide the generation towards samples correlated to information in $\mathbf{y}$. The conditional reverse process can be approximated using a Gaussian distribution, resembling the unconditional case, but with an adjusted mean:
\begin{equation}
    \begin{aligned}
        p_{\theta, \psi}(\mathbf{x}_{t-1} | \mathbf{x}_t, \mathbf{y}) \approx
        \mathcal{N} \Big( \mathbf{x}_{t-1}; \, &\mu_{\theta}(\mathbf{x}_t, t) + \Sigma_{\theta}(\mathbf{x}_t, t) \cdot \nabla_{\mathbf{x}_t} \log p_{\psi}(\mathbf{y}|\mathbf{x}_t), \\
        &\Sigma_{\theta}(\mathbf{x}_t, t) \Big).
    \end{aligned}
\label{eq:classifier_guided}
\end{equation}
For DDIMs, a score-based conditioning trick \citep{song2020denoising, song2019generative} can be applied to define a new function approximator for conditional sampling:
\begin{equation}
    \hat{\epsilon}_{\theta}(\mathbf{x}_t, t) = \epsilon_{\theta}(\mathbf{x}_t, t) - \sqrt{1 - \bar{\alpha}_t} \cdot \nabla_{\mathbf{x}_t} \log p_{\psi}(\mathbf{y}|\mathbf{x}_t).
\label{eq:guided_noise}
\end{equation}
Based on this concept the authors of \cite{pdae_zhang2022unsupervised} employed a gradient estimator \(G_\psi(\mathbf{x}_t, \mathbf{y}_{sem},t)\) to simulate \(\nabla_{\mathbf{x}_t}\log{p(\mathbf{y}_{sem}|\mathbf{x}_t})\) that assemble a conditional DPM as a decoder. The decoder is conditioned on the semantic encoder output that forces it to learn more information to improve the generation of a frozen and unconditional pretrained DDPM. More details will be discussed in Section \ref{sec:rl} since we used the same concepts but on a compressed data representation as discussed in Section \ref{sec:compression_model} and Section \ref{sec:ldm}.

\section{Methods}
\label{sec:methods}

\begin{figure*}[!ht]
\centering
\includegraphics[width=\textwidth]{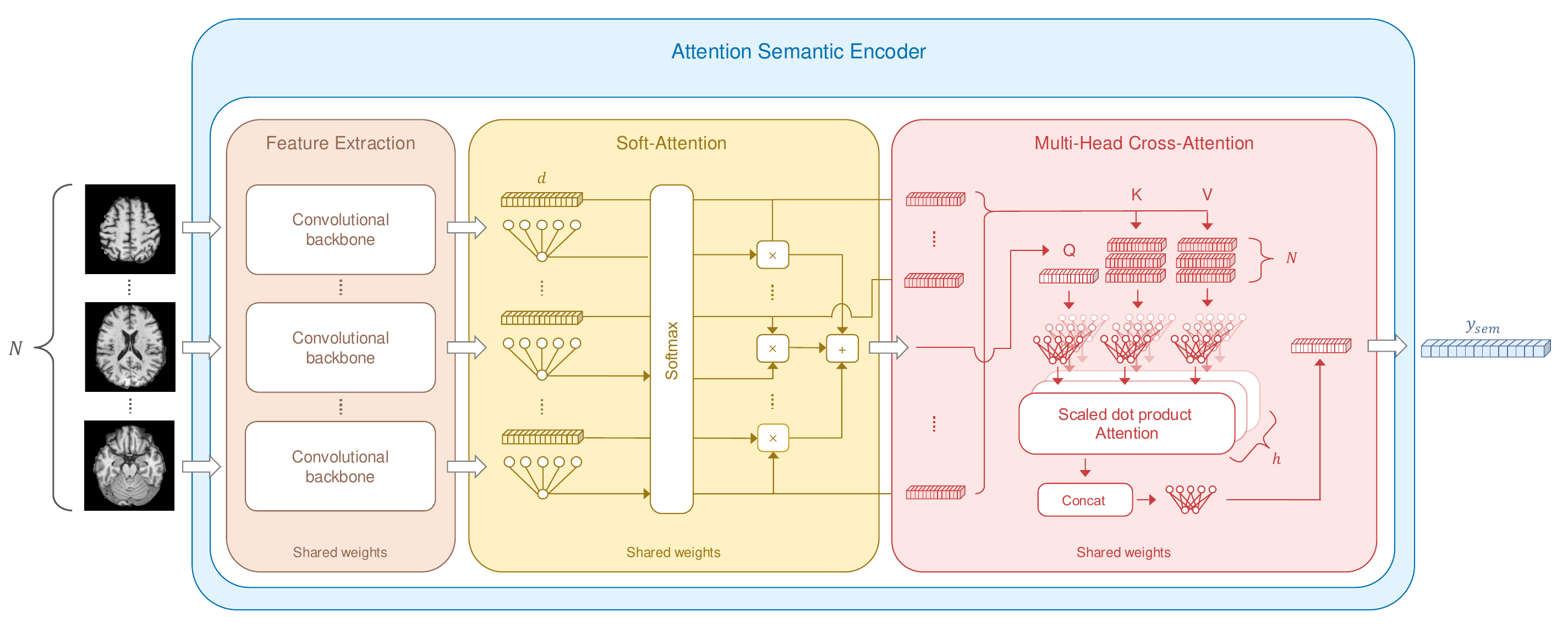}
\caption{
        \textbf{Architecture of the Semantic Encoder used in the LDAE framework.}
        The input 3D brain MRI scan $x_0 \in \mathbb{R}^{H \times W \times D}$ is sliced along the axial plane into a sequence of 2D slices, each processed independently through a shared 2D CNN backbone (e.g., ConvNeXt-Small) to extract slice-level embeddings $e_i \in \mathbb{R}^d$. These embeddings are then aggregated via a two-stage attention mechanism: \textit{SoftAttention} computes a global summary vector $Q$ as a weighted mean over the sequence, and \textit{CrossAttention} simulates self-attention by querying $Q$ with the original embeddings $E$, yielding a final global non-spatial semantic vector $y_{\text{sem}} \in \mathbb{R}^d$ used to guide the reverse diffusion process.
    }
\label{fig:semantic_encoder_design}
\end{figure*}

The overall scheme of the proposed framework is shown in Fig.\ref{fig:approach_overview}. This section provides a detailed description of the training stages and the network architecture choices. Since the framework is designed to learn two different types of latent representations from the original data, we will refer throughout the remainder of the manuscript to the latent space generated by the compression model as the \textit{compressed space} (Section \ref{sec:compression_model}) and to the latent space produced by the semantic encoder as the \textit{semantic space} (Section \ref{sec:rl}) 

\subsection{Compression Model}
\label{sec:compression_model}
To make the training tractable, given the 3D nature of the input data, we followed the principles of LDMs as already done by \cite{brain_diff_longitudinal_puglisi2024enhancing, brain_diff_pinaya2022brain}. As in the original LDMs \citep{sd_rombach2022high}, we trained an AE based on \citep{aekl_esser2021taming} as our perceptual compression model, an essential step to scale to high-resolution images. This model consists of an AE trained by a combination of a perceptual loss \citep{percep_loss_zhang2018unreasonable} and a patch-based adversarial objective \citep{adv1_dosovitskiy2016generating, adv2_yu2021vector}. These losses guarantee that reconstructions remain within the image manifold by enforcing local realism while also preventing the blurriness that arises from relying solely on pixel-space losses like L2 or L1 objectives. To avoid arbitrarily high variance in the compressed space, we used the Kullback-Leibler (KL) regularization, which imposes a slight KL penalty that encourages the compressed space to stay close to a normal distribution, similar to Variational Autoencoders (VAEs) \citep{vae_kingma2013auto, vae_rezende2014stochastic}. Hence, given a 3D brain scan \( \mathbf{x_0} \in \mathbb{R}^{H \times W \times D \times 1}\), a compression encoder \(\mathcal{E}\) encodes \(\mathbf{x_0}\) into a compressed representation \(\mathbf{z_0}=\mathcal{E}(\mathbf{x_0}) \in \mathbb{R}^{h \times w \times d \times c}\) in which \(f=H/h= W/w=D/d\) is the downsampling factor and the decoder \(\mathcal{D}\) reconstructs the image from the compression: \(\tilde{\mathbf{x}_0}=\mathcal{D}(\mathbf{z}_0)=\mathcal{D}(\mathcal{E}(\mathbf{x}_0))\). 

\subsection{Latent Diffusion Model - Pretraining}
\label{sec:ldm}
The trained perceptual compression model provides a compact representation of the input scan by a factor \(f\) along each dimension, leading to a volume with \(f^3\) times fewer voxels. This model discards imperceptible high-frequency details while retaining crucial semantics embedded in low-frequency structures, making it an efficient yet expressive space for generative modelling. However, to ensure stability and improve the effectiveness of diffusion training, we normalize the learned latents as suggested in Appendix G. of \cite{sd_rombach2022high}. Specifically, after encoding with \(\mathcal{E}\), we rescale the latent representation to have unit variance across the first batch, ensuring a standardized latent space that facilitates diffusion learning. 

In this stage, we train a DDPM without conditioning on the input image, resulting in a time-conditioned U-Net operating in the compressed space. The reweighted lower bound is:
\[L_{LDM} = \sum_{t=1}^{T} \mathbb{E}_{\mathcal{E}(x), \epsilon_t} \left[ \left\| \epsilon_{\theta} (\mathbf{z}_t, t) - \epsilon_t \right\|_2^2 \right]
\]
in which \(\mathbf{\epsilon_t} \in \mathbb{R}^{h \times w \times d \times c} \sim \mathcal{N}(\mathbf{0}, \mathbf{I})\), \(\mathbf{z_t}=\sqrt{\alpha_t}\mathbf{z_0}+\sqrt{1-\alpha_t}\mathbf{\epsilon_t}\) and \(t \sim \mathcal{U}(0,T)\) during training with \(T=1000\). The reverse DDIM process using Eq.\ref{eq:reverse_diffusion} will sample a new \(\hat{\mathbf{z_0}}\) from the compressed latent distribution \(p(\mathbf{z)}\) that can be decoded to image space trough \(\mathcal{D}\).

\subsection{Representation Learning - LDAE}
\label{sec:rl}
Now that we have pretrained the LDM we can employ a semantic encoder that maps \(\mathbf{x}_0\) into \(\mathbf{y}_{sem} = Enc_\phi(\mathbf{x}_0)\) and a gradient estimator for the compressed space guidance simulation \(G_\psi(\mathbf{z}_t, \mathbf{y}_{sem},t)\). Similarly to \cite{pdae_zhang2022unsupervised} the gradient estimator \(G_\psi\) is used to simulate \(\nabla_{\mathbf{z}_t}\log{p(\mathbf{y}_{sem}|\mathbf{z}_t})\) and the conditional decoder will approximate:
\begin{equation}
    \begin{split}
        p_{\theta, \psi}(\mathbf{z}_{t-1} | \mathbf{z}_t, \mathbf{y}_{sem}) = 
        \mathcal{N} \Big( \mathbf{z}_{t-1}; \mu_{\theta}(\mathbf{z}_t, t) + \quad \quad \quad \quad \quad \quad \\
        + \Sigma_{\theta}(\mathbf{z}_t, t) \cdot G_{\psi}(\mathbf{z}_t, \mathbf{y}_{sem}, t), \Sigma_{\theta}(\mathbf{z}_t, t) \Big).
    \end{split}
\label{eq:conditional_gaussian}
\end{equation}
LDAE is trained as a regular DPM by optimizing the following variational lower bound derived objective:
\begin{equation}
    \begin{split}
        L_{LDAE}(\psi, \phi) = \mathbb{E}_{\mathbf{x}_0, t, \epsilon} \Bigg[
        \lambda_t \Bigg\| \epsilon - \epsilon_{\theta}(\mathbf{z}_t, t)  + \quad \quad \quad \quad \quad \quad \\
        + \frac{\sqrt{\alpha_t} \sqrt{1 - \bar{\alpha}_t}}{\beta_t} 
        \cdot \Sigma_{\theta}(\mathbf{z}_t, t) \cdot G_{\psi}(\mathbf{z}_t, Enc_{\phi}(\mathbf{x}_0), t) 
        \Bigg\|^2 \Bigg].
    \end{split}
\label{eq:rl_loss_function}
\end{equation}
Note that: (i) the pretrained LDM is frozen during this phase, so Eq. \ref{eq:rl_loss_function} don't optimize parameters \(\theta\) but only \(\psi\) and \(\phi\); and (ii) the gradient estimator operates on the compressed latent distribution (\(p(\mathbf{z})\)) while the encoder on the original image (\(\mathbf{x}_0\)). This adopted optimization, similar to PDAE, forces the predicted mean shift \(\Sigma_{\theta}(\mathbf{z}_t, t) \cdot G_{\psi}(\mathbf{z}_t, Enc_{\phi}(\mathbf{x}_0), t)\) to fill the latent posterior mean gap \(\tilde{\mathbf{\mu}}_t(\mathbf{z}_t, \mathbf{z}_0) - \mu_\theta(\mathbf{z}_t,t)\) by learning as much information as possible \(\mathbf{y}_{sem}\) from \(\mathbf{x}_0\). 

Following \cite{pdae_zhang2022unsupervised}, we adopt their proposed weighting scheme of the diffusion loss (Eq. \ref{eq:rl_loss_function}), which has been shown to improve training stability and representation learning. Instead of using a constant weighting factor (\(\lambda_t = 1\)), they suggest a signal-to-noise ratio (SNR)-based scheme:
\begin{equation}
    \lambda_t = \left( \frac{1}{1 + \text{SNR}(t)} \right)^{1-\gamma} \cdot \left( \frac{\text{SNR}(t)}{1 + \text{SNR}(t)} \right)^\gamma,
\end{equation}
where \(\text{SNR}(t) = \frac{\bar{\alpha}_t}{1 - \bar{\alpha}_t}\) and \(\gamma\) is a hyperparameter controlling the balance between early-stage and late-stage weighting. Following their recommendation, we set \(\gamma = 0.1\) as it down-weights the loss for both very low and very high diffusion steps while encouraging the model to focus on learning richer representations in intermediate stages. 

\subsection{Encoder Design}
\label{sec:encoder_design}
To enable efficient training and convergence in unsupervised representation learning, we adopted an encoder design inspired by our previous diagnostic framework proposed in \cite{lozupone2024axial}. In that work, we observed that leveraging 2D CNN architectures improved convergence speed and performance when training with limited data.

Our encoder leverages a 2D CNN for feature extraction while back-propagating the error signal at the volume level. Specifically, we treat axial slices as a sequential input of 2D images, which are embedded by a 2D CNN. Formally, given a sequence of axial slices $ x = {x_1, x_2, \dots, x_L} $, the embedding function $ f_{\text{cnn}}(\cdot) $ maps each slice to a feature representation:
\[e_i = f_{\text{cnn}}(x_i), \quad e_i \in \mathbb{R}^{d}, \quad i = 1, \dots, L\]
where $ d $ is the embedding dimension. Since self-attention transforms a sequence $ {e_1, \dots, e_L} $ into another sequence and we want a non-spatial compact vector \(y_{sem}\), we approximate this mechanism through a combination of soft-attention and cross-attention mechanisms. Soft-attention computes a weighted mean representation of the sequence, where the weights indicate the relative importance of each embedding:
\[Q = \text{SoftAttention}(E), \quad E = [e_1, e_2, \dots, e_L] \in \mathbb{R}^{L \times d}\]
This yields a general representation $ Q \in \mathbb{R}^{1 \times d} $, computed as:
\[Q = \sum_{i=1}^{L} \alpha_i e_i, \quad \alpha_i = \frac{\exp(w_i)}{\sum_{j=1}^{L} \exp(w_j)}\]
where $ w_i $ are learnable attention weights. Cross-attention operation is applied to approximate self-attention one by using the global representation $ Q $ obtained from soft-attention as the query. Instead of directly computing self-attention on the full sequence, we compute:
\[\text{MultiHeadAttention}(Q, K, V)\]
where $ Q \in \mathbb{R}^{1 \times d} $ is the global summary, and both the key and value matrices are the original sequence $ K = V = E \in \mathbb{R}^{L \times d} $. The resulting single-head attention matrix has the shape:
\[
\text{Attention}(Q, K, V) = \text{softmax} \left( \frac{Q K^T}{\sqrt{d}} \right) V \in \mathbb{R}^{1 \times d}
\]
This formulation ensures that the final representation \(y_{sem}\) encapsulates the most relevant features from the sequence while maintaining global dependencies, and allows yielding state-of-the-art 2D CNN architectures as base embedding networks, with the additional flexibility to utilize pre-trained weights.

\subsection{Gradient-Estimator Design}
The gradient estimator $G_\psi(\mathbf{z}_t, \mathbf{y}_{sem}, t)$ is implemented as a modified U-Net architecture as proposed in \citep{pdae_zhang2022unsupervised}. It shares the same downsampling path and time embedding modules as the pre-trained LDM introduced in Section~\ref{sec:ldm}, enabling reuse of the learned representations from the unconditional model. However, to enable conditional guidance based on the semantic encoding $\mathbf{y}_{sem}$, we add a second, dedicated upsampling path. As a result, $G_\psi$ consists of a shared encoder and two decoder branches:
\begin{enumerate}
    \item The first branch (original) corresponds to the unconditional denoiser $\epsilon_\theta$ trained during LDM pretraining.
    \item The second branch is newly initialized, including its middle blocks, upsampling path, and output layers, and uses the same skip connections of the frozen encoder. This branch is trained from scratch using Eq. \ref{eq:rl_loss_function}.
\end{enumerate}
Following the conditioning strategy proposed in~\citep{pdae_zhang2022unsupervised}, we use AdaGN~\citep{dhariwal2021diffusion} to inject both timestep $t$ and semantic condition $\mathbf{y}_{sem}$ into the new decoder. AdaGN applies a learned affine transformation to the normalized feature maps:
\[
\text{AdaGN}(\mathbf{h}, t, \mathbf{y}_{sem}) = \mathbf{y}_{sem}^s \left( \mathbf{t}_s \cdot \text{GroupNorm}(\mathbf{h}) + \mathbf{t}_b \right) + \mathbf{y}_{sem}^b,
\]
where $[\mathbf{y}_{sem}^s, \mathbf{y}_{sem}^b]$ and $[\mathbf{t}_s, \mathbf{t}_b]$ are obtained via linear projections of $\mathbf{y}_{sem}$ and $t$, respectively. This design allows the model to retain the low-level representations learned during the unconditional pretraining, while enabling the new branch to learn how to guide the denoising process using the high-level semantic codes. 

\subsection{Gradient Estimator as Stochastic Encoder and Conditional Decoder}
\label{sec:gradient_est_enc_dec}
Beyond approximating the denoising function $\epsilon_\theta$, the gradient estimator $G_\psi(\mathbf{z}_t, \mathbf{y}_{sem}, t)$ plays a dual role in our LDAE framework: it can act as both a \emph{stochastic encoder} and a \emph{conditional decoder}. This dual capability was demonstrated in the original DAE framework \citep{dae_preechakul2022diffusion} and is essential to enable both efficient reconstruction and semantic-level control.

\paragraph{Stochastic Encoder via DDIM Inversion}
To encode a compressed sample \(\mathbf{z_0}=\mathcal{E}(\mathbf{x_0})\) into its stochastic latent code $\mathbf{z}_T$, we employ a deterministic DDIM-like backward using the semantic code $\mathbf{y}_{sem} = \text{Enc}_\phi(\mathbf{x}_0)$ and the gradient estimator \(G_\psi\):
\begin{equation}
\mathbf{z}_{t+1} = \sqrt{\bar{\alpha}_{t+1}} f_\theta(\mathbf{z}_t, t, \mathbf{y}_{sem}) + \sqrt{1 - \bar{\alpha}_{t+1}}\, \epsilon_\theta(\mathbf{z}_t, t, \mathbf{y}_{sem}),
\end{equation}
where $f_\theta(\mathbf{z}_t, t, \mathbf{y}_{sem})$ is the DDIM predictor and $\epsilon_\theta$ is approximated via the conditional gradient estimator $G_\psi$. This process encodes the residual information not captured by $\mathbf{y}_{sem}$ into $\mathbf{z}_T$, enabling near-exact reconstructions.

\paragraph{Conditional Decoder for Reconstruction and Generation}
Conversely, the same gradient estimator can decode a noisy latent code $\mathbf{z}_T$ into a clean latent $\mathbf{z}_0$ via the conditional DDIM reverse (generative) process. This decoding can be performed in two modes:
\begin{enumerate}
  \item \textbf{Reconstruction:} starting from the inferred $\mathbf{z}_T$ obtained via inversion and $\mathbf{y}_{sem}$, enabling near-exact reconstructions.
  \item \textbf{Generation:} starting from pure noise $\mathbf{z}_T \sim \mathcal{N}(0, I)$ and guiding the process with $\mathbf{y}_{sem}$, producing samples that shares semantic attributes with $x_0$.
\end{enumerate}
This dual capability allows the model to interpolate in latent space, perform counterfactual generation, and reconstruct missing intermediate timepoints, as explored in our semantic evaluation experiments (see Section~\ref{sec:experiments}).

\subsection{Controlling Semantic Attributes via Latent Linear Directions}
\label{sec:disentanglement}

Once a semantically rich encoder is trained (Sections ~\ref{sec:rl}--\ref{sec:encoder_design}), 
we can manipulate the latent representation \(\mathbf{y}_{sem} \in \mathbb{R}^d\) to alter specific 
features in the reconstructed 3D scan as proposed in \cite{dae_preechakul2022diffusion}. 
Concretely, suppose we train a linear classifier
\[
\ell(\mathbf{y}_{sem}) \;=\; \mathbf{w}^\top \,\mathbf{y}_{sem} \;+\; b
\]
to distinguish, for example, AD from CN participants. The set of points 
\(\mathbf{y}_{sem}\) satisfying \(\mathbf{w}^\top \mathbf{y}_{sem} + b = 0\) 
forms a hyperplane in \(\mathbb{R}^d\). By definition, the weight vector 
\(\mathbf{w}\) is orthogonal to this hyperplane, meaning that
\(\mathbf{w}^\top (\mathbf{y}_{sem,2} - \mathbf{y}_{sem,1}) = 0\) for any two 
points \(\mathbf{y}_{sem,1}\) and \(\mathbf{y}_{sem,2}\) on the decision boundary. 

Hence, \(\mathbf{w}\) defines a principal direction in \(\mathbf{y}_{sem}\)-space 
along which movement most strongly affects the classifier’s output 
(\(\ell(\mathbf{y}_{sem})\)). Intuitively, translating \(\mathbf{y}_{sem}\) in the 
direction of \(\mathbf{w}\) (i.e., \(\mathbf{y}_{sem} \leftarrow \mathbf{y}_{sem} + \alpha \mathbf{w}\)) 
will “add” the corresponding AD-related features to the reconstructed scan, 
whereas moving in the opposite direction (\(\alpha < 0\)) will “subtract” them. 
Empirically, this directional manipulation can disentangle specific symptoms or 
morphological traits in the semantic representation, thereby giving control 
over generated 3D reconstructions.

\subsection{Semantically meaningful interpolation}
\label{sec:interpolation}
To assess the semantic smoothness of the learned semantic space, we performed interpolation leveraging both the semantic and stochastic codes. This enables qualitative and quantitative evaluation of how well the latent space captures continuous and clinically meaningful transformations across brain MRI scans.

Following the DAE framework \cite{dae_preechakul2022diffusion}, we perform linear interpolation in the semantic space and spherical linear interpolation in the stochastic one. Given two input scans $\mathbf{x}_1$ and $\mathbf{x}_2$, their corresponding semantic and stochastic representations are denoted as $\mathbf{y}_{\text{sem}}^1, \mathbf{z}_T^1$ and $\mathbf{y}_{\text{sem}}^2, \mathbf{z}_T^2$, respectively. The interpolated latent representations at interpolation factor $t \in [0, 1]$ are computed as:

\begin{flalign}
    \text{LERP}(\mathbf{y}_{\text{sem}}^1, \mathbf{y}_{\text{sem}}^2; t) &= (1 - t) \cdot \mathbf{y}_{\text{sem}}^1 + t \cdot \mathbf{y}_{\text{sem}}^2 \label{eq:lerp} \\
    \text{SLERP}(\mathbf{z}_T^1, \mathbf{z}_T^2; t) &= \frac{\sin((1 - t) \theta)}{\sin(\theta)} \mathbf{z}_T^1 + \frac{\sin(t \theta)}{\sin(\theta)} \mathbf{z}_T^2 \label{eq:slerp}
\end{flalign}
where the angle $\theta$ between $\mathbf{z}_T^1$ and $\mathbf{z}_T^2$ is given by:
\begin{equation}
    \theta = \arccos \left( \frac{\langle \mathbf{z}_T^1, \mathbf{z}_T^2 \rangle}{\|\mathbf{z}_T^1\| \cdot \|\mathbf{z}_T^2\|} \right)
\end{equation}
The interpolated pair $(\mathbf{y}_{\text{sem}}^{(t)}, \mathbf{z}_T^{(t)})$ is then decoded using the reconstruction procedure described in Section~\ref{sec:gradient_est_enc_dec}.

\begin{figure*}[ht]
    \centering
    \includegraphics[width=1.0\textwidth]{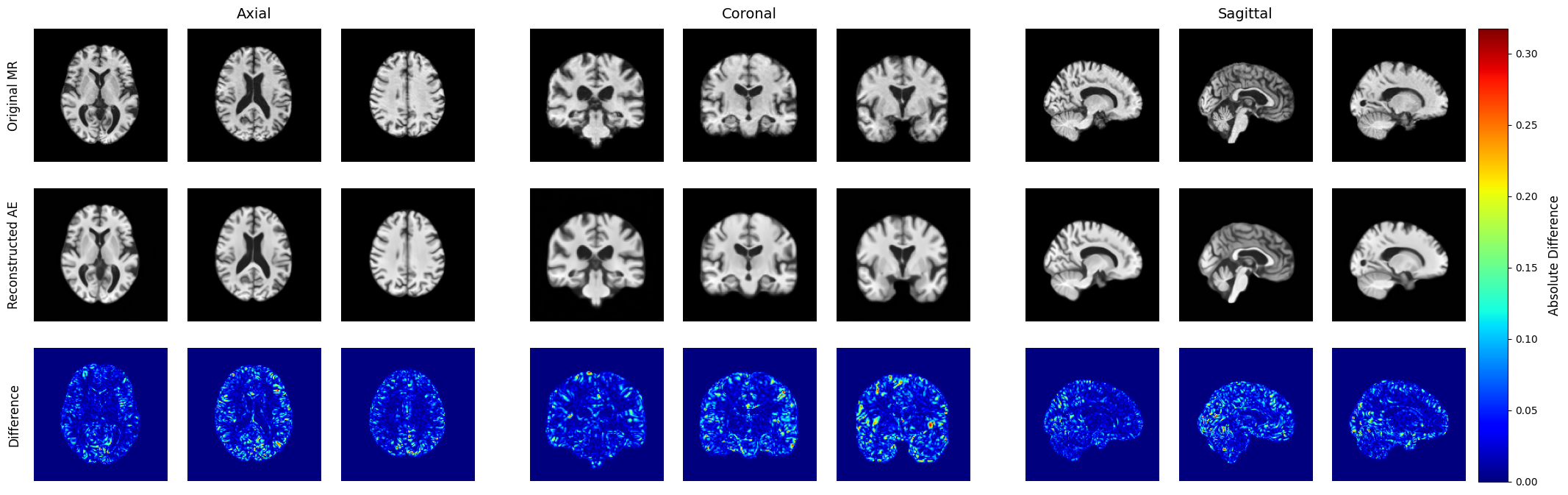}
    \caption{Qualitative reconstructions from AutoencoderKL. Top row: original scan slices. Middle row: reconstructed outputs from compressed latent codes. Bottom row: reconstruction error. The reconstructions preserve global and local anatomical features despite the $170\times$ compression.}
    \label{fig:ae_recon}
\end{figure*}

\begin{figure*}[ht]
    \centering
    \includegraphics[width=0.49\textwidth]{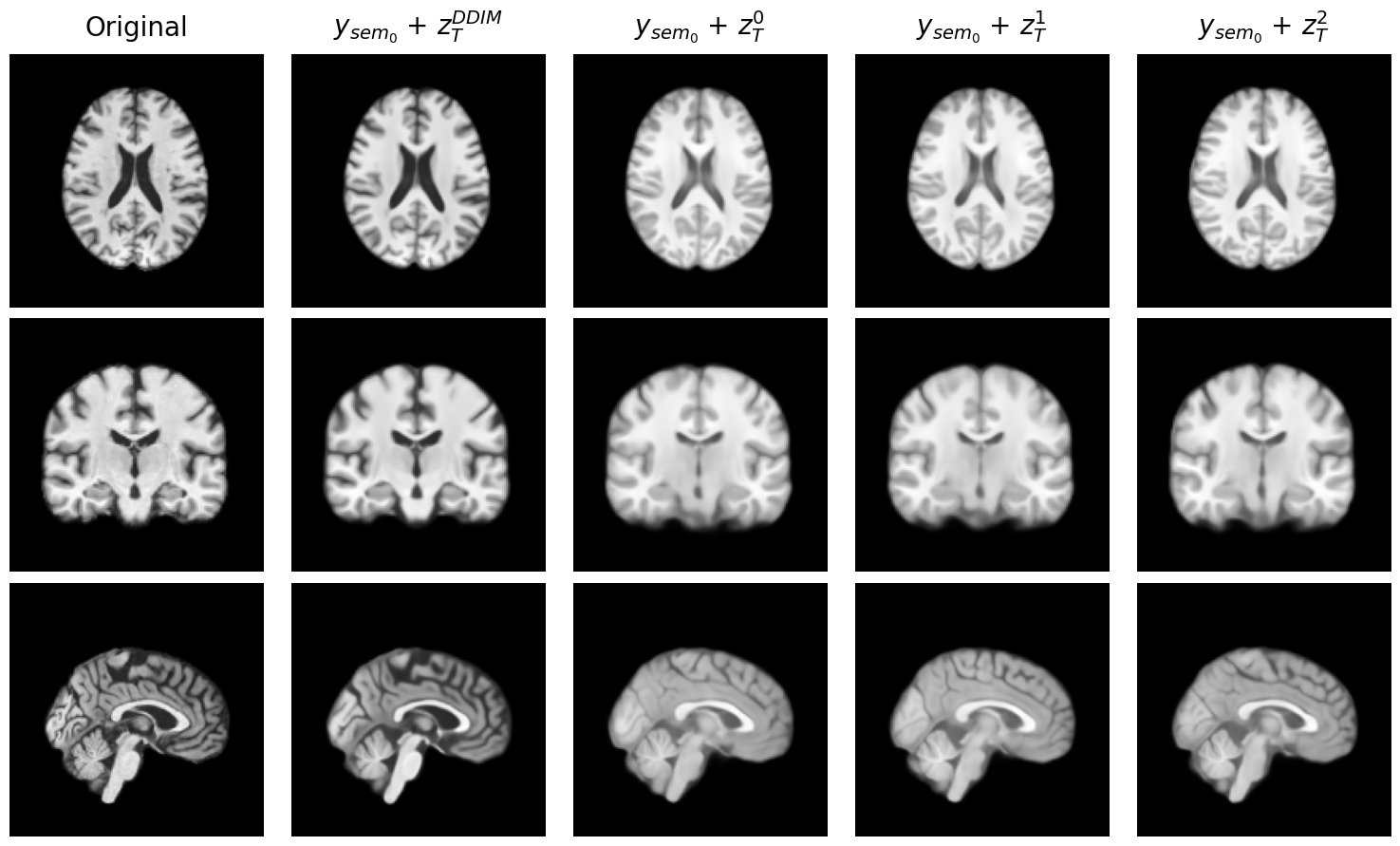} \hspace{0.011\textwidth}
    \includegraphics[width=0.49\textwidth]{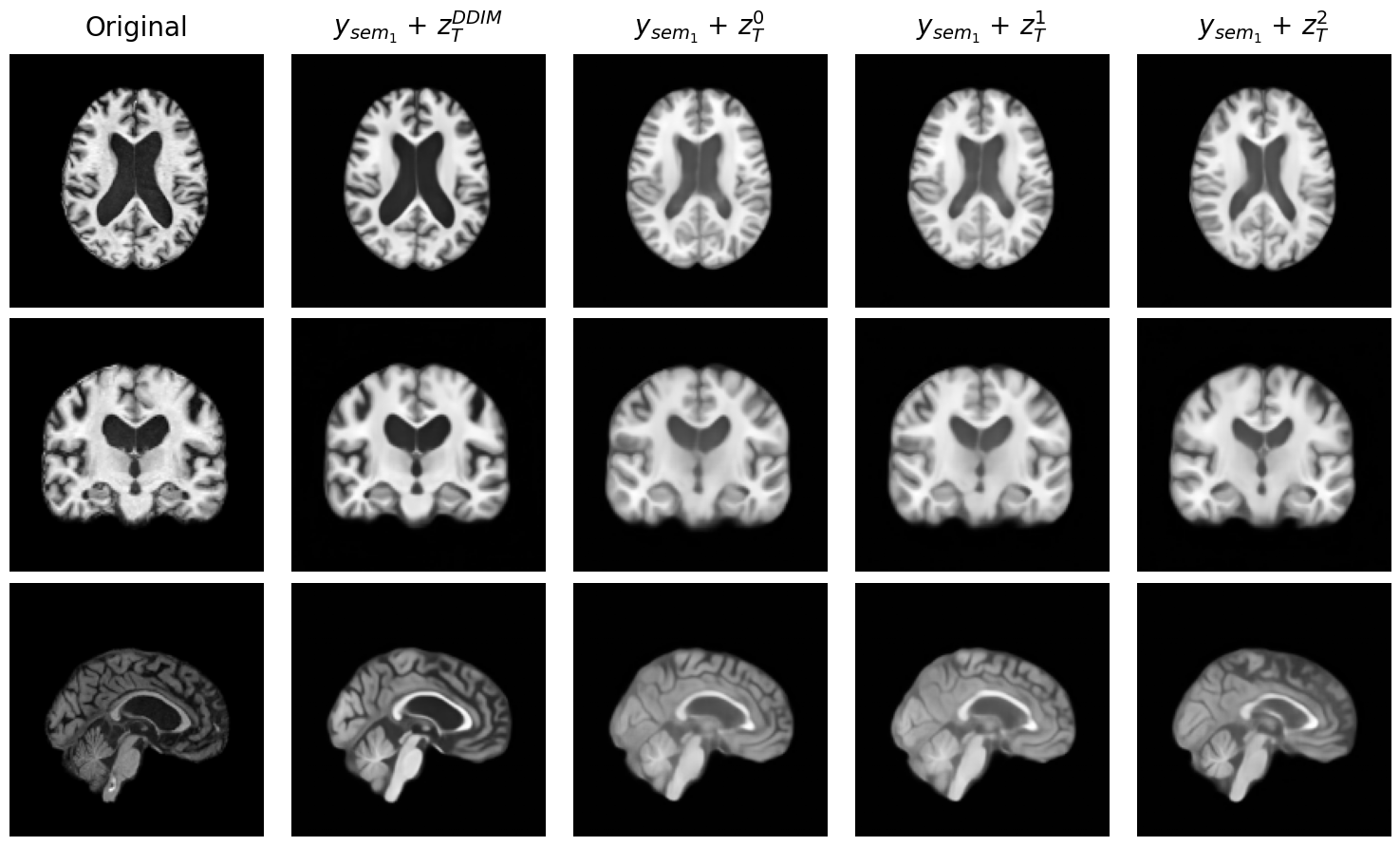}
    \caption{
        \textbf{Reconstruction from semantic code and stochastic latent.} Reconstruction results for two representative subjects from the test set: a CN subject (left block) and an AD subject (right block). 
        \textbf{First column:} original brain MR scan.
        \textbf{Second column:} reconstruction obtained using both the semantic embedding \( y_{\text{sem}} = \text{Enc}_\phi(x_0) \) and the stochastic latent \( z_T \) obtained via DDIM inversion of the encoded compressed latent \( z_0 = \mathcal{E}(x_0) \).
        \textbf{Columns 3–5:} reconstructions obtained by keeping \( y_{\text{sem}} \) fixed and sampling different \( z_T^{(i)} \sim \mathcal{N}(0, I) \). Despite the stochastic variation, the reconstructions retain global anatomical and disease-relevant structure, indicating that \( y_{\text{sem}} \) captures high-level semantics while \( z_T \) encodes low-level variability. In the AD subject, expected pathological traits (e.g., ventricular enlargement) are consistently preserved, whereas in the CN subject, normal cortical volume and structure remain stable across samples.
    }
    \label{fig:ldae_recon}
\end{figure*}

\section{Experiments}
\label{sec:experiments}
\subsection{Dataset and Preprocessing}

We conduct our experiments on the ADNI database containing longitudinal 3D brain volumes of subjects across various stages of cognitive decline, ranging from cognitively normal (CN) to mild cognitive impairment (MCI) and Alzheimer’s disease (AD). The subjects' demographic information of the dataset used is reported in Table \ref{tab:demographic}, which provides an essential context for our analysis and findings.

\paragraph{BIDS Conversion}  
We first converted the raw ADNI data into the Brain Imaging Data Structure (BIDS) format \citep{gorgolewski2016brain}. This conversion enables structured and standardized neuroimaging data handling and seamless integration with Python-based neuroimaging tools such as PyBIDS \citep{yarkoni2019pybids}. We performed this conversion using the Clinica platform \citep{routier2021clinica, samper2018reproducible}, an open-source software framework specifically designed for reproducible clinical neuroscience research. During this phase, the Clinica ADNI-to-BIDS converter automatically applies a quality control check, selecting the preferred scan for each visit and discarding scans that fail predefined quality checks.

\bgroup
\def\arraystretch{1.5}
\begin{table}[!t]
    \caption{Population statistics across CN, AD, and MCI groups. Including age, mini-mental state examination (MMSE) and global clinical dementia rating (CDR) scores}
    \label{tab:demographic}
    \centering
    \begin{adjustbox}{width=\columnwidth}
    \begin{tabular}{cccccc}
    \hline
    & Subjects & Samples & Age & MMSE & CDR \\  \hline
    CN & 965  & 3673  & $75.43 \pm 6.83$ & $29.06 \pm 1.20$ & $0.03 \pm 0.17$ \\ 
    AD & 748  & 2064  & $76.36 \pm 7.59$ & $21.79 \pm 4.44$ & $0.95 \pm 0.52$ \\
    MCI & 1193 & 4603  & $74.72 \pm 7.74$ & $27.51 \pm 2.24$ & $0.46 \pm 0.22$ \\
    \hline
    
    \end{tabular}
\end{adjustbox}
\end{table}
\egroup

\paragraph{Preprocessing Pipeline}  
The preprocessed dataset was prepared using a standard neuroimaging pipeline composed of the following sequential steps:

\begin{enumerate}
    \item \textbf{Bias Field Correction:} Intensity non-uniformities were corrected using the N4ITK algorithm \citep{tustison2010n4itk}.
    \item \textbf{Skull Stripping:} Brain tissue was extracted using the deep learning-based brain extraction models proposed in \cite{cullen2018convolutional} from ANTsPyNet package.
    \item \textbf{Affine Registration:} Each brain-extracted volume was affinely aligned to the MNI152 ICBM 2009c nonlinear symmetric template \citep{fonov2009unbiased, fonov2011unbiased} using the SyN algorithm from the ANTs toolkit \citep{avants2008symmetric, avants2014insight}.
\end{enumerate}

\paragraph{Inference-Time Normalization}  
We employed the MONAI framework \citep{cardoso2022monai} for batch preprocessing at inference time. The following transformations were applied:

\begin{enumerate}
    \item \textbf{Voxel Spacing Normalization:} All images were resampled to a uniform voxel spacing of 1.5mm isotropic using B-spline interpolation.
    \item \textbf{Spatial Resizing:} Volumes were resized using cropping or padding to a target shape (e.g., $128 \times 160 \times 128$).
    \item \textbf{Intensity Normalization:} Intensities were rescaled to the $[0, 1]$ range using min-max normalization.
\end{enumerate}

\subsection{Experimental Setup}
\paragraph{AutoencoderKL Training}

To enable efficient diffusion training in a compressed representation space, we fine-tuned a 3D perceptual autoencoder to compress full-resolution MRI scans into a compact latent representation. We used the \texttt{AutoencoderKL} implementation from the \texttt{MONAI Generative} \citep{pinaya2023generative} framework. The model was initialized from a checkpoint pre-trained on the UK Biobank dataset \citep{sudlow2015uk} and fine-tuned on volumes resampled to a resolution of $128 \times 160 \times 128$. The training setup is summarized in Table~\ref{tab:autoenc_hyperparams}.

\begin{table}[h]
\captionsetup{justification=raggedright,singlelinecheck=false}
\caption{AutoencoderKL architecture and training configuration.}
\resizebox{\columnwidth}{!}{%
\begin{tabular}{ll}
\toprule
\textbf{Parameter} & \\
\midrule
Input Resolution & $1 \times 128 \times 160 \times 128$ \\
Latent Size & $3 \times 16 \times 20 \times 16$ \\
Channels & [64, 128, 128, 128] \\
Residual Blocks per Level & 2 \\
Normalization & Group Normalization \\
KL Weight & $1 \times 10^{-7}$ \\
Adversarial Weight & 0.025 \\
Perceptual Weight & 0.001 \\
Perceptual Net & SqueezeNet (fake-3D ratio: 0.5) \\
Discriminator Layers / Channels & 3 / 64 \\
Optimizer & Adam \\
Learning Rate (Generator) & $5 \times 10^{-5}$ \\
Learning Rate (Discriminator) & $1 \times 10^{-4}$ \\
Batch Size (Effective) & 8 (1 GPUs x 1, grad. accum. 8) \\
Training Time & 20 epochs / $\sim$ 3 days \\
Pretrained Init & UK Biobank \citep{brain_diff_pinaya2022brain} \\
Hardware & 1 $\times$ NVIDIA A100 80GB \\
\bottomrule
\end{tabular}
} 
\label{tab:autoenc_hyperparams}
\end{table}

\paragraph{Latent Diffusion Model Pretraining}

Following the compression model training stage, we pretrained a DDPM directly in the compressed latent space. This stage models the distribution of latent representations $z_0 \in \mathbb{R}^{3 \times 16 \times 20 \times 16}$, significantly reducing the computational burden compared to operating in voxel space. The training follows the standard DDPM formulation with a linear noise schedule and uses a 3D UNet architecture trained to predict the added Gaussian noise in the latent space. The Exponential Moving Average (EMA) of model weights update was applied during training for improved stability and sample quality. A full list of architectural and training parameters is provided in Table~\ref{tab:ldm_pretraining}.

\begin{table}[h]
\captionsetup{justification=raggedright,singlelinecheck=false}
\caption{Latent Diffusion Model (LDM) pretraining configuration}
\resizebox{\columnwidth}{!}{%
\begin{tabular}{ll}
\toprule
\textbf{Parameter} & \\
\midrule
Input Latent Shape & $3 \times 16 \times 20 \times 16$ \\
Channels & [256, 512, 768] \\
Residual Blocks per Level & 2 \\
Attention Resolutions (factors) & [2, 4] \\
Dropout & 0.1 \\
Timesteps & 1,000 \\
Beta Schedule & Linear: $\beta_t \in [10^{-4}, 2 \times 10^{-2}]$ \\
Optimizer & Adam \\
Learning Rate & $2.5 \times 10^{-5}$ \\
EMA Decay & 0.999 \\
Batch Size (Effective) & 128 (2 GPUs × 64, grad. accum. = 2) \\
Training Time & 500 epochs / $\sim$ 18 hours \\
Hardware & 2 × NVIDIA A100 SXM4 40GB  \\
\bottomrule
\end{tabular}
} 
\label{tab:ldm_pretraining}
\end{table}

\begin{table}[h]
\captionsetup{justification=raggedright,singlelinecheck=false}
\caption{Representation Learning Stage: Configuration of the Semantic Encoder and Gradient Estimator}
\resizebox{\columnwidth}{!}{%
\begin{tabular}{ll}
\toprule
\multicolumn{2}{l}{\textbf{Semantic Encoder}} \\
\midrule
Architecture & 2.5D Attention-based Encoder \\
Slicing plane & Axial \\
Backbone & \texttt{ConvNeXt-Small} \\
Pretrained Init & ImageNet \\
Input Modality & $1 \times 128 \times 128 \times 160 $ \\
Input Sequence Length & 128 slices \\
Multihead Attention & 8 heads, attention dropout 0.1 \\
Output Representation & Global, non-spatial vector $\mathbf{y}_{sem} \in \mathbb{R}^{768}$ \\
\midrule
\multicolumn{2}{l}{\textbf{Gradient Estimator $G_\psi$}} \\
\midrule
Architecture & Modified U-Net (same configuration of LDM) \\
Shared Components & Downsampling blocks from pretrained LDM \\
New Components & Middle, upsampling, and output blocks \\
Condition Injection & AdaGN \\
Weighting Scheme & SNR based, $\gamma = 0.1$ \\
\midrule
\multicolumn{2}{l}{\textbf{Training Configuration}} \\
\midrule
Optimizer & Adam \\
Learning Rate & $2.5 \times 10^{-5}$ \\
EMA Decay & 0.999 \\
Effective Batch Size & 16 (2 GPUs × 2 × grad. accum. 4) \\
Training Duration & 200 epochs / $\sim$2 days \\
Hardware & 2 × NVIDIA A100 SXM4 40GB \\
\bottomrule
\end{tabular}
}
\label{tab:rep_learning_config}
\end{table}

\paragraph{Representation Learning via 2.5D Semantic Encoder}

To guide the reverse diffusion process with clinically meaningful features, we trained a semantic encoder network to extract a compact latent code $\mathbf{y}_{sem}$ from the original 3D brain volume $\mathbf{x}_0$. This latent code serves as a conditioning vector during denoising via the gradient estimator $G\psi(\mathbf{z}_t, \mathbf{y}_{sem}, t)$, as detailed in Section~\ref{sec:rl}. The encoder follows a 2.5D strategy, where axial slices are independently processed using a 2D CNN backbone, and then aggregated using a combination of soft attention and cross-attention mechanisms to produce a global, non-spatial embedding. We employed a pre-trained ConvNeXt-Small model as the slice-level feature extractor, adapted to accept single-channel inputs (grayscale MRI slices), with embedding dimension $d=768$ and input sequence length $L=128$. The grayscale adaptation was done by summing the
pre-trained convolutional filters of the backbone’s first layer. The semantic encoder and gradient estimator were optimized jointly to minimize the weighted diffusion loss described in Equation~\ref{eq:rl_loss_function}, using the SNR-based weighting scheme proposed by Zhang et al.~\citep{pdae_zhang2022unsupervised}. The gradient estimator $G_\psi$ was implemented as a modified U-Net architecture, reusing the encoder and time embedding modules from the pre-trained unconditional LDM (Table~\ref{tab:ldm_pretraining}). A new decoder branch, composed of middle, upsampling and output blocks, was initialized with the same configuration of the LDM.

A full list of architecture and training parameters for this stage is provided in Table~\ref{tab:rep_learning_config}.

\begin{table*}[ht!]
\footnotesize
\captionsetup{justification=raggedright,singlelinecheck=false}
\caption{Comparison of voxel-space DAE and latent-space LDAE in terms of input resolution, model size, training duration, and reconstruction efficiency.}
\resizebox{\textwidth}{!}{%
\begin{tabular}{lccccccc}
\toprule
\textbf{Model} & \textbf{Input Resolution} & \textbf{Model Parameters} & \textbf{Training Duration} & \textbf{Hardware} & \textbf{Inference Time ($T=100$)} & \textbf{SSIM (↑) ($T=100$)} & \textbf{LPIPS (↓) ($T=100$)} \\
\midrule
DAE & $112 \times 128 \times 112$ & 130M & 140 epochs / $\sim$1 week & 2 × A100 40GB & $\sim$120 seconds & 0.892 & \textbf{0.038} \\
LDAE & $128 \times 160 \times 128$ & 920M & 500+200 epochs / $\sim$3 days & 2 × A100 40GB & $\sim$6 seconds & \textbf{0.962} & 0.076 \\
\bottomrule
\end{tabular}
}
\label{tab:efficiency_comparison}
\end{table*}

\begin{table*}[h]
\centering
\caption{Comparison of  \textbf{SSIM}, \textbf{LPIPS}, and \textbf{MSE} for various models at different sampling steps $\mathbf{T}$. Since training a full-resolution \textbf{3D Diffusion Autoencoder (DAE)} was computationally prohibitive, we resized the scans to \textbf{1x112×128×112}. However, we trained the \textbf{AutoencoderKL} at higher resolution \textbf{1x128×160×128}, allowing it to encode images into a \textbf{3x16×20×16} latent space. This enabled efficient training of \textbf{LDDIM} and \textbf{LDAE}, which operate in the compressed latent space but rely on the autoencoder for final image reconstruction. As a result, AutoencoderKL reconstruction quality acts as a bottleneck for LDDIM and LDAE performance.}
\resizebox{\linewidth}{!}{
\begin{tabular}{lrcccccccccccc}
\toprule
\multirow{2}{*}{\textbf{Model}} & \multirow{2}{*}{\textbf{Latent dim}} & 
\multicolumn{4}{c}{\textbf{SSIM} (↑)} & \multicolumn{4}{c}{\textbf{LPIPS} (↓)} & \multicolumn{4}{c}{\textbf{MSE} (↓)} \\
\cmidrule(lr){3-6}\cmidrule(lr){7-10}\cmidrule(lr){11-14}
&  & \textbf{T=10} & \textbf{T=20} & \textbf{T=50} & \textbf{T=100} &
\textbf{T=10} & \textbf{T=20} & \textbf{T=50} & \textbf{T=100} &
\textbf{T=10} & \textbf{T=20} & \textbf{T=50} & \textbf{T=100} \\
\midrule

\textbf{AutoencoderKL (@14M)} & 5,120
 & \multicolumn{4}{c}{0.962} 
 & \multicolumn{4}{c}{0.075}
 & \multicolumn{4}{c}{0.001} \\

 \midrule

\textbf{LDDIM (@486M)} & & & & & & & & & & & & & \\
a) Encoded $x_T$ 
 & 5,120
 & 0.959 & 0.962 & 0.962 & \textbf{0.962}
 & 0.077 & 0.075 & 0.075 & 0.075
 & \textbf{0.001} & \textbf{0.001} & \textbf{0.001} & \textbf{0.001} \\

\midrule
\textbf{LDAE (@920M)} & & & & & & & & & & & & & \\
a) No encoded $x_T$, from $y_{sem}$
 & 768
 & 0.872 & 0.871 & 0.870 & 0.869 
 & 0.156 & 0.155 & 0.155 & 0.155 
 & 0.005 & 0.005 & 0.005 & 0.005 \\
b) Encoded $x_T$ 
 & 5,888
 & 0.953 & 0.960 & 0.962 & \textbf{0.962} 
 & 0.081 & 0.077 & 0.076 & 0.075
 & \textbf{0.001} & \textbf{0.001} & \textbf{0.001} & \textbf{0.001} \\

\midrule
\textbf{DAE (@130M)} & & & & & & & & & & & & & \\
a) No encoded $x_T$, from $y_{sem}$
 & 768
 & 0.272 & 0.287 & 0.281 & 0.283
 & 0.460 & 0.256 & 0.182 & 0.170
 & 0.017 & 0.016 & 0.016 & 0.016 \\
b) Encoded $x_T$ 
 & 1,605,632
 & 0.216 & 0.397 & 0.684 & 0.892
 & 0.433 & 0.132 & 0.038 & \textbf{0.030}
 & 0.007 & 0.002 & \textbf{0.001} & \textbf{0.001} \\

\bottomrule
\end{tabular}
}
\label{tab:diffae_comparison}
\end{table*}

\paragraph{Data Splits and Model Selection Strategy}

We adopted a consistent 90-10 train-test split at the subject level for all training stages. Within the 90\% training partition, a random 1\% subset was reserved for validation purposes. This minimal validation split was chosen to preserve maximal training data availability, as the main objective was image reconstruction rather than classification. Validation was conducted using image-level reconstruction metrics. For the \textit{compression model} and the \textit{representation learning stage}, model selection was based on the best SSIM score observed on the validation set. In the representation learning stage, we specifically monitored the SSIM between the original image and the reconstruction generated using the semantic embedding $\mathbf{y}_{sem}$ and pure noise as stochastic input. This metric served as a proxy for the quality and completeness of the semantic information extracted by the encoder. No validation-based checkpointing was used in the \textit{latent diffusion model pretraining} stage. Instead, the model parameters at the final epoch were retained for downstream use, as unconditional diffusion sampling was the primary goal, and model weights updates was performed trough an EMA strategy.

\paragraph{Linear Probe Experimental Setup}

To evaluate the semantic quality of the learned representations, we performed linear probe experiments on two downstream tasks: AD diagnosis and age prediction. For this purpose, we trained linear models on top of the fixed semantic embeddings extracted from the trained LDAE encoder. Given an input 3D brain MRI scan $x_0$, we first projected each image into the semantic space using the encoder $\text{Enc}_{\phi}(x_0)$, obtaining a semantic vector $y_{\text{sem}} \in \mathbb{R}^{768}$. These embeddings were computed for all samples in the dataset. The dataset was initially split at the subject level into 90\% training and 10\% test sets. From the 90\% training portion, we held out 1\% for validation and used the remaining 89\% for training. We further split this 89\% into 70\% training and 30\% validation subsets for the linear probe experiments. The original 10\% test set was kept as a fixed benchmark for final evaluation. For the AD vs. CN classification task, we trained a linear classifier consisting of a single fully connected layer with binary cross-entropy loss. We trained a linear regressor using mean squared error (MSE) loss for the age prediction task.  The classifier was trained using the Adam optimizer for 200 epochs with a $1 \times 10^{-3}$ learning rate. The regressor was trained using stochastic gradient descent (SGD) for 1000 epochs with the same learning rate. Note that all linear models were trained on the fixed semantic vectors without finetuning the encoder.

These linear probe experiments aim to quantify the extent to which the learned semantic codes $y_{\text{sem}}$ encode clinically meaningful and linearly separable features relevant to disease classification and age estimation.

\paragraph{Interpolation Experiment for Missing Scan Generation}

We conducted a semantic and stochastic interpolation experiment on a subset of the held-out test set to evaluate the model's ability to generate plausible missing follow-up scans. This experiment simulates longitudinal scan prediction by reconstructing intermediate brain scans from a subject's early and late visits. For each subject with at least three visits in the test set, we selected all valid $(start, target, end)$ triplets such that the target timepoint lies temporally between the start and end. For each triplet, we computed the interpolation factor $\alpha$ based on the temporal offset of the target with respect to the interval defined by start and end. We then interpolated linearly between the semantic codes ($y_{\text{sem}}$) and spherically between the stochastic codes ($z_T$) to generate the latent pair $(y_{\text{sem}}^{(t)}, z_T^{(t)})$ corresponding to the target intermediate timepoint $t$. This experiment was conducted on a subset of 30 subjects from the test set, yielding approximately $1400$ valid triplet configurations. The number of possible triplets $T$ for a subject with $n$ sessions grows approximately as $T \propto \binom{n}{3}$, under the constraint that the target scan lies strictly between the start and end. This implies that even modest increases in subjects considered for the evaluation will increase significantly the number of generations to compute. The generated latent representation was decoded using the DDIM-based reverse process described in Section \ref{sec:interpolation}. The resulting image $\hat{x}_0$ was compared to the ground truth scan $x_0^{\text{target}}$ using SSIM and MSE. We report average SSIM and MSE metrics stratified by the temporal gap between start and end scans (time gap), the minimum distance of the predicted target from the endpoints (prediction gap), and the relative position of the target within the interval (normalized to $[0,1]$). This allows us to assess how interpolation accuracy varies with temporal context.

\section{Results}
\label{sec:results}

In this section, we present experimental results evaluating the proposed LDAE framework. The results are structured to progressively validate the components of the pipeline and support the two main hypotheses introduced in Section~\ref{sec:intro}.

\subsection{AutoencoderKL: Perceptual Compression and Reconstruction Quality}
We begin by evaluating the perceptual compression model based on \texttt{AutoencoderKL}. This autoencoder compresses each 3D brain MR from $1 \times 128 \times 160 \times 128$ into a latent representation of size $3 \times 16 \times 20 \times 16$, reducing the volume by a factor of approximately 170$\times$. Despite this compression, the model achieves high-fidelity reconstructions, with an SSIM of 0.962 and an MSE of 0.001 on the external test set (see Table~\ref{tab:diffae_comparison}). Qualitative examples are shown in Figure~\ref{fig:ae_recon}. These results highlight the model's capacity to retain high-frequency anatomical details in the compressed latent space.

It is important to note that this reconstruction accuracy establishes an upper bound for the performance of the subsequent latent diffusion models (LDDIM and LDAE), as the final decoded output always passes through the AutoencoderKL decoder.

\subsection{Reconstruction Quality}

As shown in Figure \ref{fig:ldae_recon} and Table \ref{tab:diffae_comparison}, the proposed LDAE---when using both the semantic encoder and gradient estimator $G_\psi$ as described in Section~3.6---achieves reconstruction quality on par with AutoencoderKL and LDDIM. For instance, at $T = 50$, LDAE obtains SSIM = 0.962, LPIPS = 0.076, and MSE = 0.001, matching the performance of both the LDDIM and the upper bound imposed by the AutoencoderKL.  Since LDAE operates in the compressed latent space, it allows efficient model parameters scalability. Specifically, the full LDAE model has approximately 920M parameters, compared to 130M of the voxel-space DAE. Despite this, LDAE remains more efficient (see Table \ref{tab:efficiency_comparison}). LDAE was trained at higher resolution ($128 \times 160 \times 128$) for 200 epochs over 2 days on 2 $\times$ A100-SXM4-40GB GPUs. In contrast, the DAE required 1 week of training on the same hardware and had to operate on downsampled volumes ($112 \times 128 \times 112$) due to memory constraints, limiting its capacity and overall performance. At inference time, the efficiency gap is even more pronounced: LDAE requires approximately 6 seconds per reconstruction at $T = 100$ steps on an A100 GPU, while the full-resolution DAE takes approximately 2 minute per scan due to the lack of compression and increased I/O overhead.

These results validate our second hypothesis: LDAE enables high-fidelity 3D brain MRI reconstruction with full semantic controllability and significantly improved computational efficiency, both during training and inference.

\subsection{Semantic Guidance Enables Reconstruction from Pure Noise}

To assess the semantic richness of the learned representation, we perform reconstructions using only the semantic code \( y_{\text{sem}} = \text{Enc}_\phi(x_0) \) and a randomly sampled stochastic code \( z_T \sim \mathcal{N}(0, I) \). This setup evaluates whether the semantic code alone can guide the reverse diffusion process to generate anatomically plausible MR that contains structure information of $x_0$.

Figure~\ref{fig:ldae_recon} shows qualitative reconstructions for two randomly selected subjects: a CN individual (left) and an AD patient (right). For both cases, reconstructions are generated by fixing the semantic code \( y_{\text{sem}} = \text{Enc}_\phi(x_0) \) and sampling multiple stochastic codes \( z_T \). Across samples, the reconstructions preserve global brain morphology, indicating that the semantic code captures the high-level anatomical and pathological attributes of the subject, while the stochastic component contributes only to low-level variability.

In the CN subject, cortical thickness and brain volume are preserved across stochastic samples. In contrast, the AD subject reconstructions consistently display expected atrophic patterns, such as enlarged ventricles and reduced hippocampal volume, despite variation in image details. These observations suggest that \( y_{\text{sem}} \) encodes disease-relevant structural information.

This behavior is consistent with prior findings in DAE ~\cite{dae_preechakul2022diffusion}, where the semantic representation governs identity and structure, and the stochastic latent controls fine-grained appearance.

\subsection{Linear Probe Evaluation on Alzheimer’s Disease Classification and Age Prediction}

\begin{figure}[!ht]
\centering
\includegraphics[width=\columnwidth]{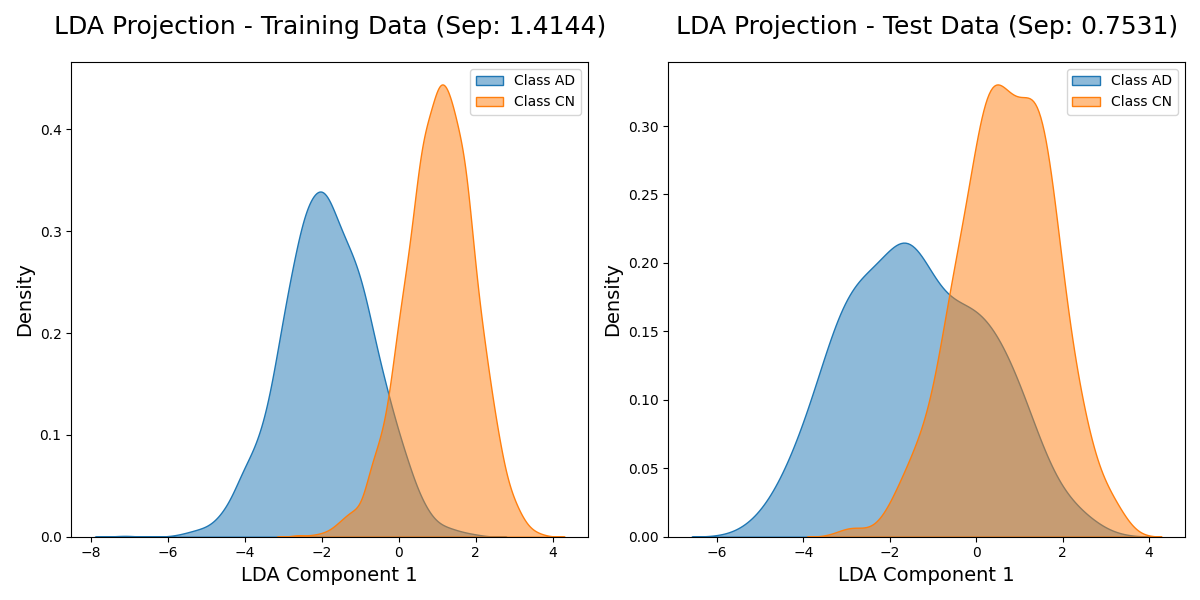}
\caption{LDA projection of semantic representations ($\mathbf{y}_{sem}$) extracted from the LDAE encoder. Each point corresponds to a 3D brain scan colored by diagnostic class (AD or CN). The clear separation suggests that the learned semantic space captures clinically meaningful features relevant to Alzheimer’s disease.}
\label{fig:lda_plot}
\end{figure}

\begin{table*}[h]
\captionsetup{justification=raggedright,singlelinecheck=false}
\caption{Linear probe evaluation using semantic representations learned by the encoders of DAE and LDAE compared to the same encoder trained with supervision. Metrics are reported separately for Alzheimer's disease classification (left) and age prediction (right).}
\footnotesize
\begin{tabularx}{\textwidth}{lXXXXXX|XX}
\toprule
\multirow{2}{*}{\textbf{Model}} 
& \multicolumn{6}{c|}{\textbf{AD vs. CN}} 
& \multicolumn{2}{c}{\textbf{Age Prediction}} \\
\cmidrule(lr){2-7} \cmidrule(lr){8-9}
& Accuracy & Precision & Recall & F1-score & MCC & ROC AUC & MAE ↓ & RMSE ↓ \\
\midrule
LDAE (Ours) & 0.8365 & 0.8469 & \textbf{0.9102} & \textbf{0.8774} & 0.6369 & 0.8948 & \textbf{4.16} & 5.23 \\
DAE (Baseline) & 0.7468 & 0.7768 & 0.8504 & 0.8119 & 0.4312 & 0.7800 & 4.93 & 6.11 \\
Supervised Encoder & \textbf{0.8464} & \textbf{0.8690} & 0.8743 & 0.8716 & \textbf{0.6806} & \textbf{0.9067} & 4.34 & \textbf{4.63} \\
\bottomrule
\end{tabularx}
\label{tab:linear_probe_joint}
\end{table*}

\begin{figure}[h!]
    \centering
    \includegraphics[width=\columnwidth]{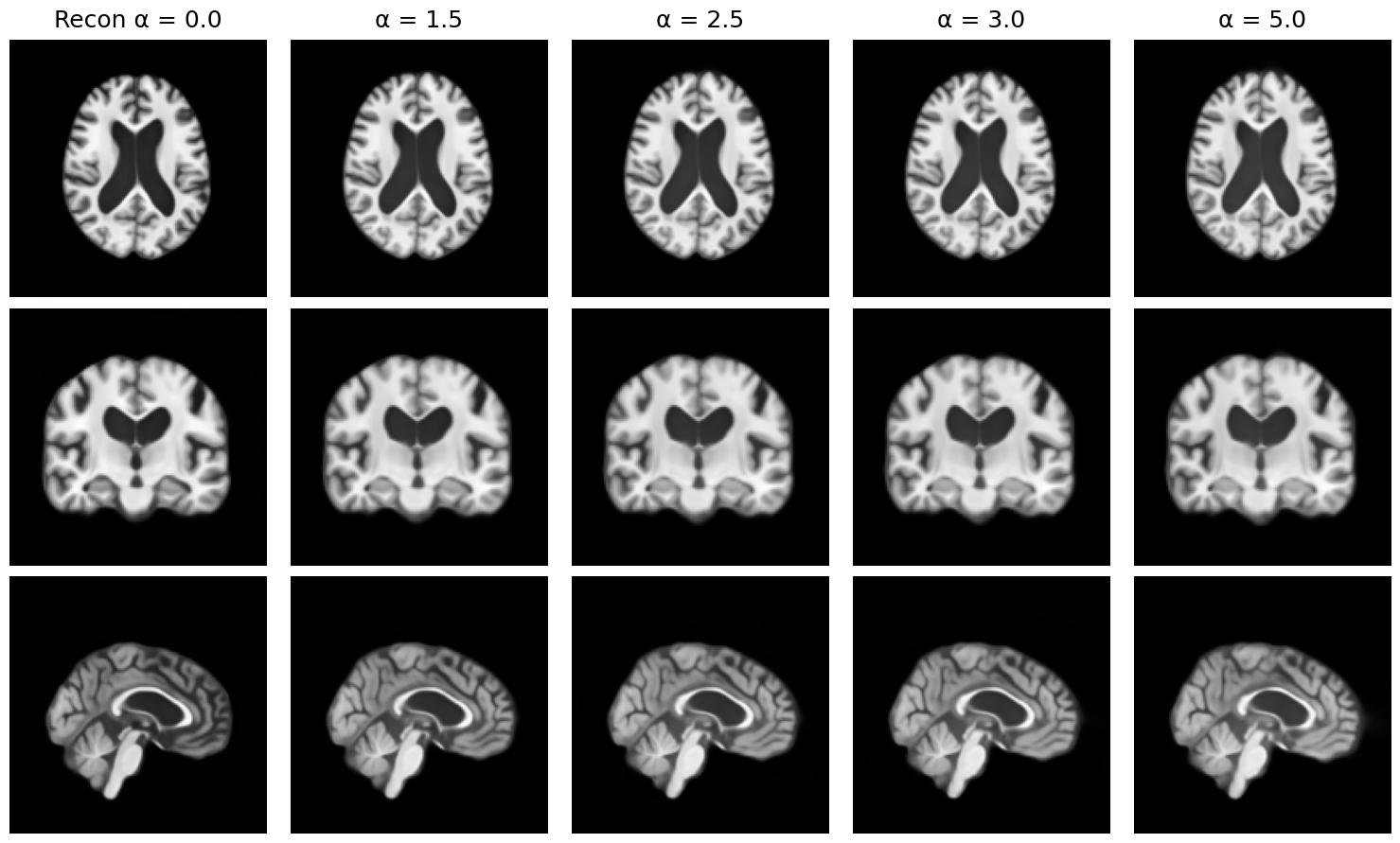}
    \caption{Progressive manipulation of an AD subject toward the CN class. The manipulation strength $\alpha$ ranges from 0.0 (original reconstruction) to 5.0. Structural changes—especially hippocampal recovery and ventricle shrinkage—become more evident with larger $\alpha$.}
    \label{fig:semantic-manipulation-scale}
\end{figure}

\begin{figure*}[ht]
    \centering
    \includegraphics[width=1.0\textwidth]{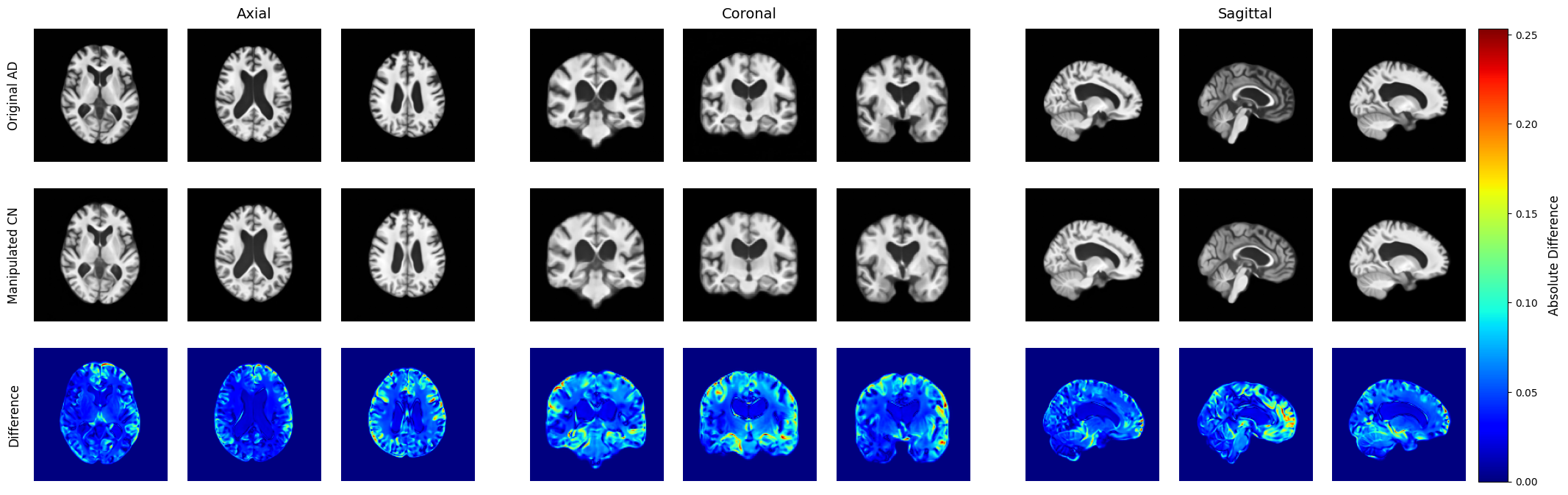}
    \includegraphics[width=1.0\textwidth]{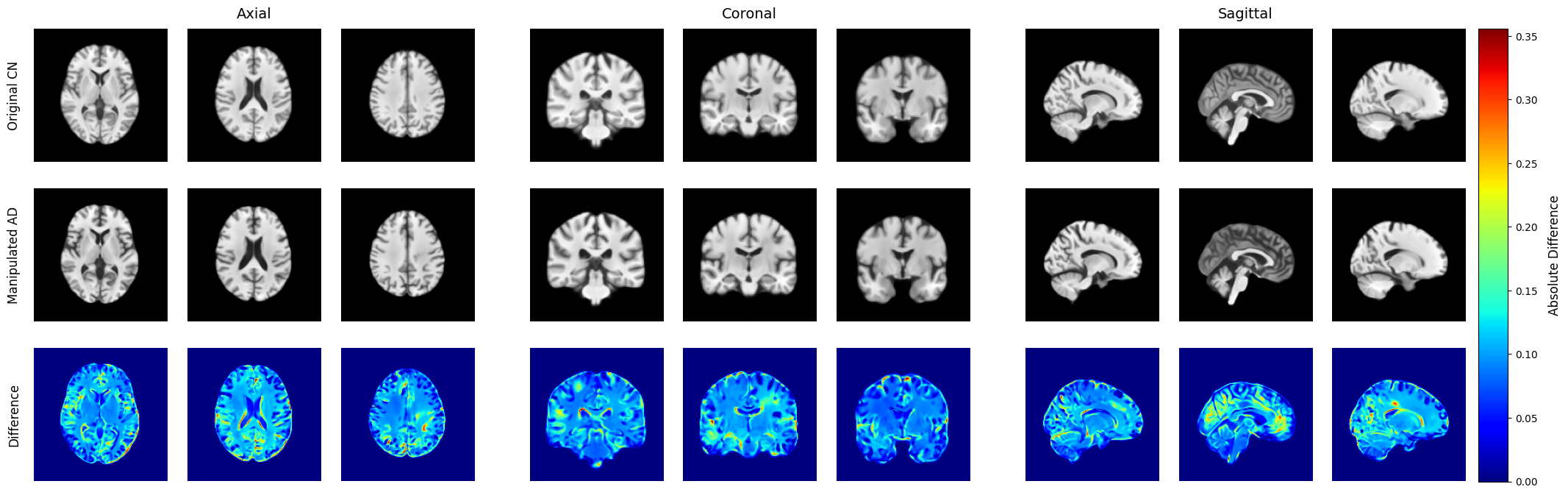}
    \caption{Semantic manipulation examples along the direction defined by the vector orthogonal to the classifier's decision boundary. 
    In the AD$\rightarrow$CN case (top), we observe a reduction of hippocampal atrophy; 
    in the CN$\rightarrow$AD case (bottom), atrophy becomes more prominent.}
    \label{fig:semantic-manipulation}
\end{figure*}

\begin{figure*}[htbp]
  \centering
  \begin{subfigure}[t]{1.0\columnwidth}
    \centering
    \includegraphics[width=\columnwidth]{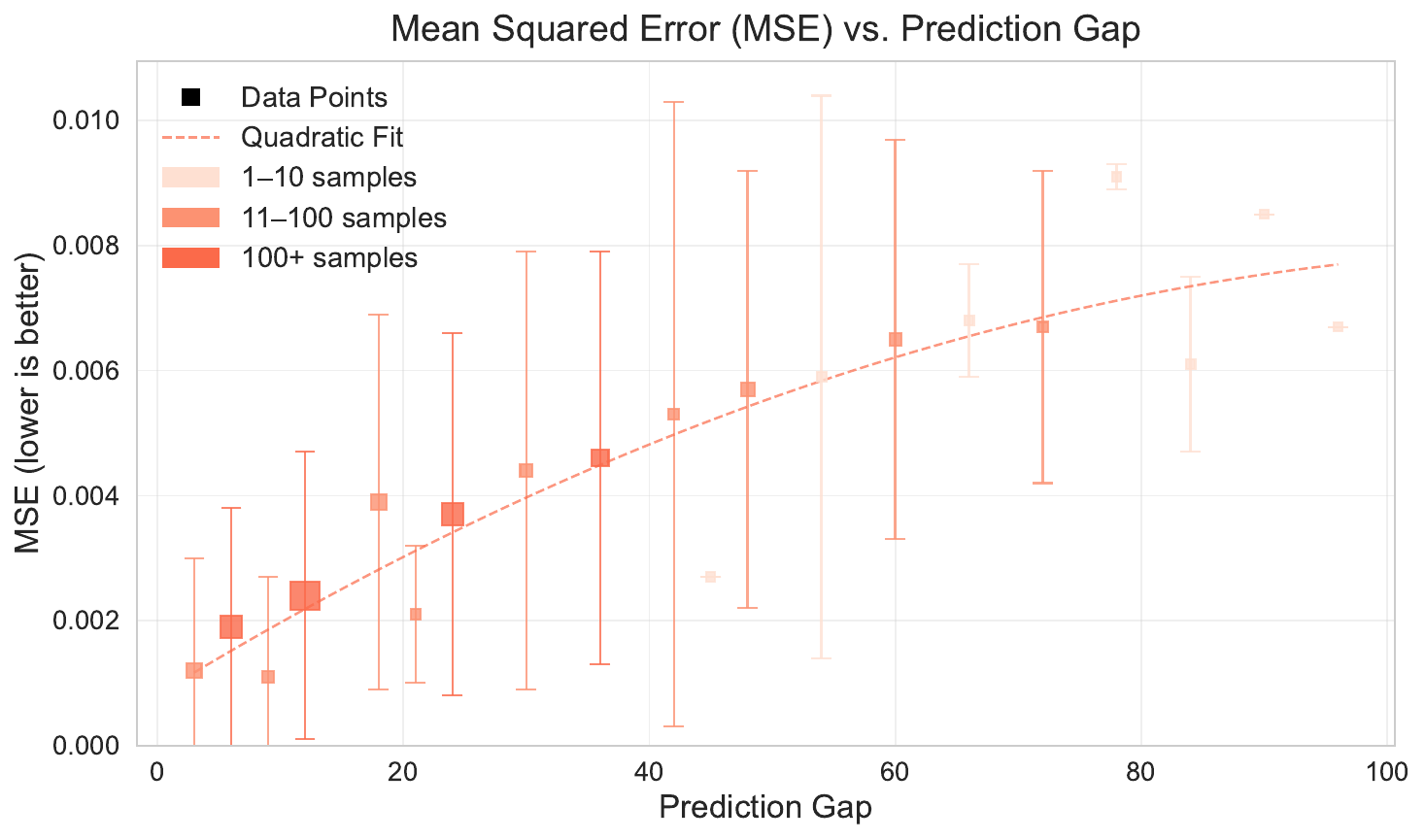}
    \caption{Mean Squared Error (MSE) as a function of prediction gap. Larger gaps correspond to more difficult interpolation tasks.}
    \label{fig:mse_predgap}
  \end{subfigure}
  \hfill
  \begin{subfigure}[t]{1.0\columnwidth}
    \centering
    \includegraphics[width=\columnwidth]{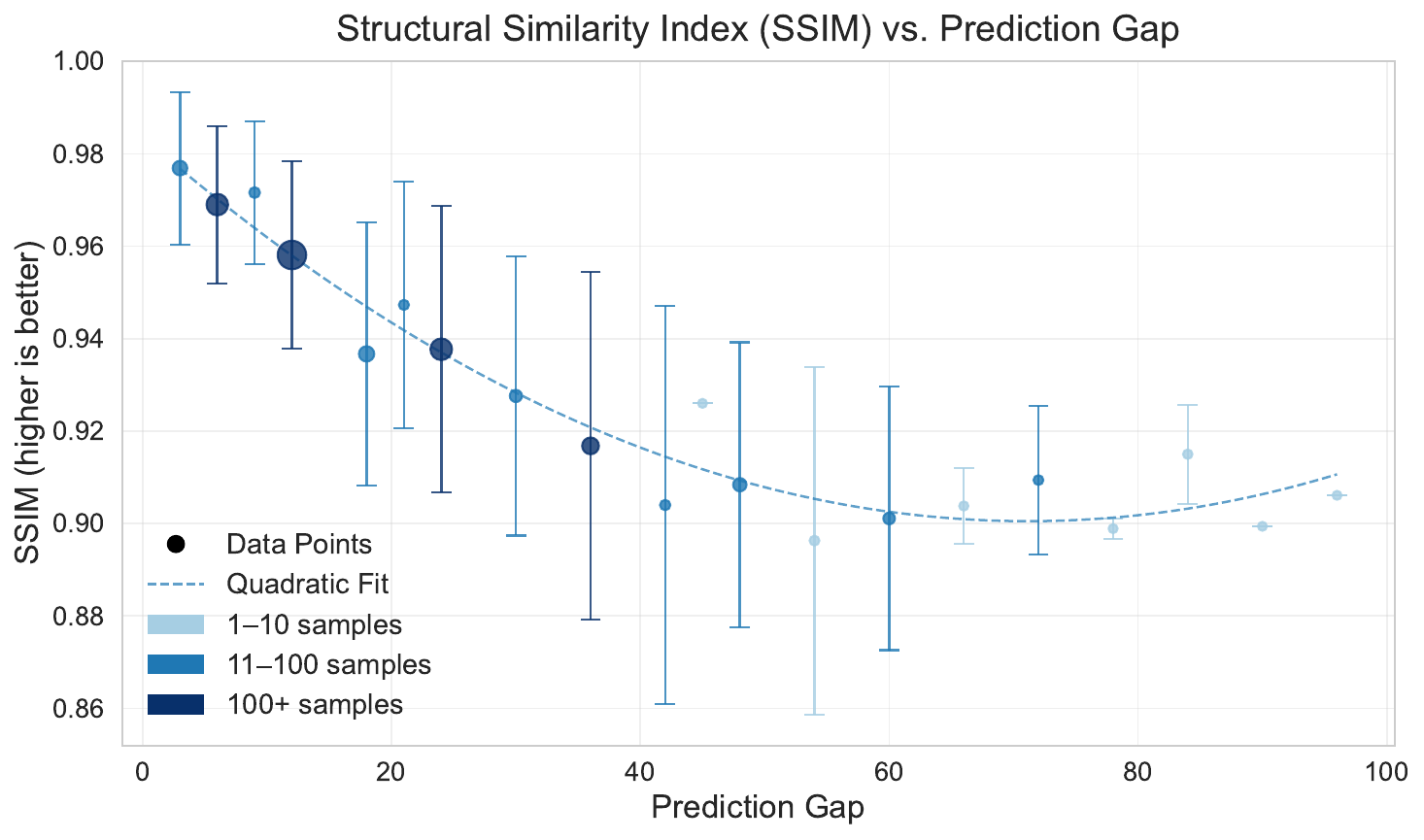}
    \caption{Structural Similarity Index (SSIM) as a function of prediction gap. The model shows robust perceptual consistency across all temporal ranges.}
    \label{fig:ssim_predgap}
  \end{subfigure}
  \caption{Quantitative evaluation of semantic interpolation across different prediction gaps. While MSE increases with the temporal distance between input scans, SSIM remains high, indicating perceptual fidelity even in challenging scenarios.}
  \label{fig:interpolation_metrics}
\end{figure*}

To quantitatively assess the semantic quality of the representations learned by our LDAE framework, we conducted linear probe experiments on two downstream tasks: (i) AD vs. CN classification, and (ii) age prediction. For this purpose, we trained a linear classifier (for AD vs. CN) and a linear regressor (for age) on top of the semantic vectors ($\mathbf{y}_{sem}$) extracted from the pre-trained LDAE encoder. As baselines, we also evaluated (i) a baseline DAE trained in the original voxel space with joint optimization of encoder and diffusion decoder and (ii) a fully supervised semantic encoder trained end-to-end with access to ground-truth diagnostic labels.

As shown in Table~\ref{tab:linear_probe_joint}, the LDAE embeddings achieved good performance on both tasks, with 83.65\% accuracy and 89.48\% AUC on the AD classification task, and a mean absolute error (MAE) of 4.16 years and RMSE of 5.23 years on the age prediction task. To further understand the semantic structure of the learned embeddings, we applied Linear Discriminant Analysis (LDA) to project the 768-dimensional semantic vectors into 2D. As shown in Figure~\ref{fig:lda_plot}, the resulting plot exhibits distinct clustering of AD and CN subjects, despite the encoder never being trained with diagnostic labels. This confirms that the semantic space $\mathbf{y}_{sem}$ captures disease-relevant information in a linearly separable manner.

These findings support Hypothesis 1, demonstrating that the semantic encoder learns semantically rich representations. Moreover, the effectiveness of such representations on both classification and regression tasks confirms their generality across clinically relevant phenotypes.

\subsubsection{Semantic Manipulation via Latent Directions}

To qualitatively evaluate the interpretability and controllability of the learned semantic space, we conducted a semantic manipulation experiment following the strategy discussed in Section \ref{sec:disentanglement}. Once a linear classifier is trained to distinguish between AD and CN subjects using the semantic representations $\mathbf{y}_{\text{sem}}$, its weight vector $\mathbf{w}$ defines the principal direction along which the most discriminative semantic variation lies. Since the classifier is trained with label 0 assigned to AD and 1 to CN, moving along $-\alpha \mathbf{w}$ amplifies AD-related features (manipulating CN towards AD). In contrast, moving in the opposite direction $\alpha \mathbf{w}$ reduces them (manipulating AD towards CN). Given a subject’s scan, we perform the manipulation as follows:
\begin{enumerate}
    \item Extract the semantic vector $\mathbf{y}_{\text{sem}} = \text{Enc}_\phi(x_0)$.
    \item Encode the stochastic latent representation $\mathbf{z}_T$ using LDAE's DDIM inversion conditioned on $\mathbf{y}_{\text{sem}}$.
    \item Perform attribute manipulation by modifying the semantic code: $\mathbf{y}_{\text{manip}} = \mathbf{y}_{\text{sem}} + \alpha \mathbf{w}$ (or $-\alpha \mathbf{w}$ depending on direction).
    \item Reconstruct the manipulated image by decoding from $(\mathbf{z}_T, \mathbf{y}_{\text{manip}})$ via the LDAE reverse process.
\end{enumerate}

Figure~\ref{fig:semantic-manipulation-scale} shows the effect of different manipulation strength, we progressively increased $\alpha$ from 0.0 to 5.0 for an AD subject. The anatomical changes become increasingly pronounced as $\alpha$ grows, particularly in hippocampal and ventricular regions, highlighting a smooth and meaningful trajectory in the latent space.
Figure~\ref{fig:semantic-manipulation} shows two representative examples: in the first case (top), an AD subject is manipulated along the $\alpha \mathbf{w}$ direction towards a CN-like representation, and in the second (bottom), a CN subject is manipulated along the $-\alpha \mathbf{w}$ direction towards an AD-like representation. Both manipulations use a scaling factor $\alpha = 1.5$.

\subsection{Interpolation for Missing Scan Generation}

To further validate the semantic consistency and interpolation capability of the learned latent spaces, we conducted interpolation experiments simulating missing follow-up scan generation. This setup shows LDAE's application toward solving the problem of reconstructing longitudinal scans that were not acquired in real studies, a frequent issue in medical datasets. This experiment leverages the ability of the LDAE framework to interpolate between a subject’s earlier and later visits to predict an intermediate scan. 

\begin{figure}[h!]
    \centering
    \includegraphics[width=\columnwidth]{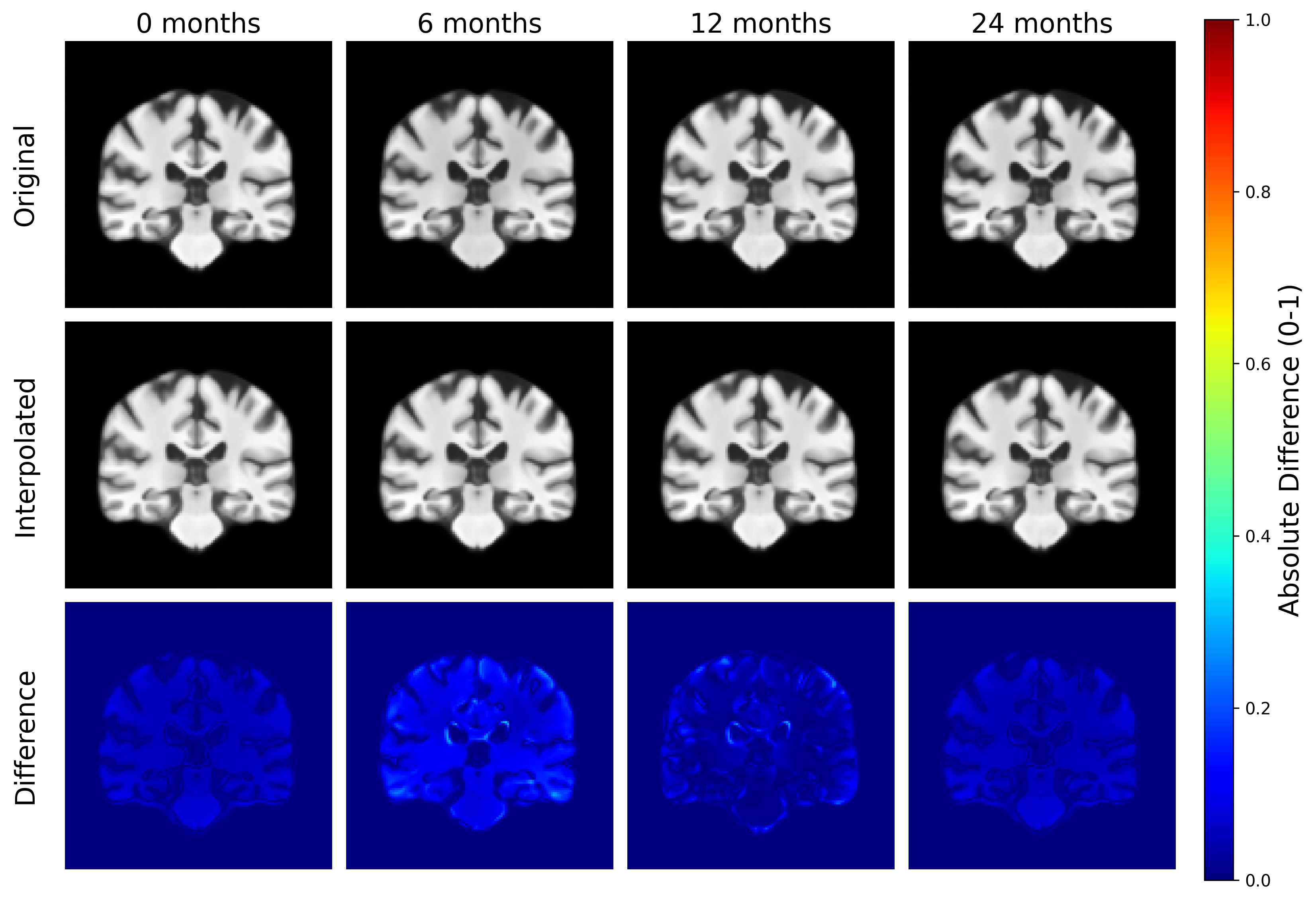}
    \caption{Qualitative example of latent interpolation for missing scan generation on a single subject with four longitudinal scans acquired at 0, 6, 12, and 24 months. The images at months 0 and 24 serve as endpoints ($\alpha = 0$ and $\alpha = 1$) for interpolation in the latent space. Intermediate scans at 6 months ($\alpha = 0.25$) and 12 months ($\alpha = 0.5$) are synthesized via linear interpolation in the semantic space and spherical interpolation in the stochastic space.}
    \label{fig:missing_values_interpolation}
\end{figure}

To quantitatively assess interpolation accuracy, we compare the generated scan $\hat{x}^{\text{target}}_0$ against the autoencoder reconstructed scan $\mathcal{D}(\mathcal{E}(x^{\text{target}}_0))$ using SSIM and MSE. This ensures the evaluation is performed entirely within the autoencoder’s output space, isolating the quality of interpolation in the compressed latent spaces from the reconstruction upperbound imposed by the compression model . The difficulty of this task is evaluated based on the \emph{prediction gap}, defined as the minimum temporal distance between the target scan and its two neighbors: $\min(\text{target} - \text{start}, \text{end} - \text{target})$. A smaller prediction gap implies a target temporally close to either neighbor (easier), while a larger gap makes accurate interpolation more challenging.

We report the results over approximately 1400 triplet configurations across 30 test subjects. As shown in Figure~\ref{fig:mse_predgap} and Figure~\ref{fig:ssim_predgap}, the LDAE maintains strong generation performance even for wider prediction gaps. Notably, SSIM remains above $0.93$ and MSE below $0.004$ even at larger temporal gaps (up to 24 months), highlighting the semantic smoothness and temporal awareness encoded in the learned representations.

In addition, Figure~\ref{fig:missing_values_interpolation} provides qualitative examples of interpolated scans across different temporal gaps. The generated images appear visually coherent and anatomically plausible, preserving the subject identity and global brain structure. These results further support the hypothesis that LDAE captures a semantically meaningful representation and potentially a temporal progression trend.

\section{Discussion and conclusion}
\label{sec:conclusions}

In this study, we introduced LDAE, a novel diffusion-based architecture specifically tailored for efficient and meaningful unsupervised representation learning, with a focused application on brain MRI scans related to AD. 

\paragraph{Semantic Representation Learning}
The results demonstrate that LDAE effectively captures structural brain changes on brain MR related to AD and aging. The linear-probe evaluations quantitatively confirm semantic representation capabilities. Although the generative results have been qualitatively validated by clinical expert, future works should include a validation of the proposed methods in clinical studies. A clinical validation is essential and can be effectively realized through close collaboration with clinical experts. Such partnerships ensure that the methodologies align with real-world medical practices, meet clinical needs, and ultimately enhance their applicability and reliability in healthcare settings. Moreover, systematic multi-attribute disentanglement analyses are essential to ensure robust and interpretable attribute manipulation, given potential correlations (e.g., disease state and age).

\paragraph{Generative Capabilities}
Regarding generative capabilities, the compressed latent space enables efficient and high-quality reconstruction. Nonetheless, reconstruction quality is inherently limited by the perceptual compression autoencoder's fidelity. It will be crucial to achieve a lossless reconstruction on a full resolution brain scan in order to use LDAE for missing scan generation and manipulation in clinical practice. Additionally, the current interpolation methods (LERP and SLERP) presume linearity in brain evolution trajectories. Incorporating advanced regression or forecasting models trained explicitly within the semantic space could better approximate realistic brain aging progressions.


\paragraph{Potential Clinical Impact}
In longitudinal studies, LDAEs could be used to capture and model subtle temporal changes of anatomical structures on medical imaging; the method can be applied for evaluating the progression of neurodegenerative diseases, or tumor growth, over time. This capability is critical for understanding disease trajectories, monitoring therapeutic responses, and personalizing treatment plans. Additionally, LDAEs are valuable in augmenting sparse datasets, a common challenge in medical research. By generating realistic synthetic data that preserves the underlying distribution of the original dataset, LDAEs can help train machine learning models more effectively. This is particularly beneficial in rare disease studies, where obtaining large datasets is often infeasible. Another significant application lies in enhancing explainability through semantic latent manipulation. By isolating and controlling specific features within the latent space, researchers and clinicians can better understand the relationships between imaging patterns and disease characteristics. 

\paragraph{General limitations and Future Work}
The success of LDAE frameworks in medical imaging depends on rigorous validations across diverse populations, imaging scanners, and disease phenotypes. Variability in imaging protocols, demographic factors, and disease presentations can significantly impact model performance. Therefore, comprehensive cross-validation is essential to ensure the generalizability and robustness of these frameworks in real-world clinical settings.

\paragraph{Foundation Model capability} 
These findings position LDAE as a promising framework for scalable and interpretable medical imaging applications and as a potential Foundation Model for 3D medical image analysis. Future work should explore its transferability to other tasks and modalities, investigate domain adaptation strategies, and benchmark its performance in diverse clinical settings to validate its generalization capacity and pretraining utility.

\section*{Declaration of generative AI and AI-assisted technologies in the writing process}
During the preparation of this work the authors used ChatGPT in order to improve language and readability. After using this tool/service, the authors reviewed and edited the content as needed and take full responsibility for the content of the publication.

\section*{Acknowledgements}
Project ECS 0000024 “Ecosistema dell’innovazione - Rome Technopole” financed by EU in NextGenerationEU plan through MUR Decree n. 1051 23.06.2022 PNRR Missione 4 Componente 2 Investimento 1.5 - CUP H33C22000420001.
\\

Data collection and sharing for the Alzheimer's Disease Neuroimaging Initiative (ADNI) is funded by the National Institute on Aging (National Institutes of Health Grant U19AG024904). The grantee organization is the Northern California Institute for Research and Education. In the past, ADNI has also received funding from the National Institute of Biomedical Imaging and Bioengineering, the Canadian Institutes of Health Research, and private sector contributions through the Foundation for the National Institutes of Health (FNIH) including generous contributions from the following: AbbVie, Alzheimer’s Association; Alzheimer’s Drug Discovery Foundation; Araclon Biotech; BioClinica, Inc.; Biogen; Bristol-Myers Squibb Company; CereSpir, Inc.; Cogstate; Eisai Inc.; Elan Pharmaceuticals, Inc.; Eli Lilly and Company; EuroImmun; F. Hoffmann-La Roche Ltd and its affiliated company Genentech, Inc.; Fujirebio; GE Healthcare; IXICO Ltd.; Janssen Alzheimer Immunotherapy Research \& Development, LLC.; Johnson \& Johnson Pharmaceutical Research \& Development LLC.; Lumosity; Lundbeck; Merck \& Co., Inc.; Meso Scale Diagnostics, LLC.; NeuroRx Research; Neurotrack Technologies; Novartis Pharmaceuticals Corporation; Pfizer Inc.; Piramal Imaging; Servier; Takeda Pharmaceutical Company; and Transition Therapeutics.

\bibliographystyle{elsarticle-harv.bst} \biboptions{comma,sort,comma,authoryear}
\bibliography{refs}




\end{document}